\documentclass[10pt,twocolumn,letterpaper]{article}

\usepackage[pagenumbers]{cvpr} % To force page numbers, e.g. for an arXiv version

\usepackage{algorithm}
\usepackage{algpseudocode}
\usepackage{makecell}
\usepackage{minted}
\usepackage[listings]{tcolorbox}
\tcbuselibrary{minted,skins}
\usepackage{xcolor}
\usepackage{stfloats}
\usepackage{balance}  % 在导言区添加
\usepackage[table]{xcolor}
\usepackage{booktabs}
\definecolor{cvprblue}{rgb}{0.21,0.49,0.74}
\usepackage[pagebackref,breaklinks,colorlinks,allcolors=cvprblue]{hyperref}

\def\paperID{12176} % *** Enter the Paper ID here
\def\confName{CVPR}
\def\confYear{2026}

\title{MVP: Multiple View Prediction Improves GUI Grounding}

\author{
Yunzhu Zhang$^{1}$\quad
Zeyu Pan$^{2}$\quad
Zhengwen Zeng$^{3}$\quad
Shuheng Shen$^{3}$$^*$\quad
Changhua Meng$^{3}$\quad
Linchao Zhu$^{1}$\thanks{Corresponding Author.}
\\
$^{1}$College of Computer Science and Technology, Zhejiang University \\
$^{2}$College of Computer Science and Technology, Hangzhou Dianzi University \\
$^{3}$Venus Team, Ant Group \\
{\tt\small \{yunzhuzhang0918,zhulinchao7\}@gmail.com, panzeyucs@hdu.edu.cn,} \\
{\tt\small \{zengzhengwen.zzw,shuheng.ssh,changhua.mch\}@antgroup.com} \\
}

\begin{document}
\maketitle
\begin{abstract}
GUI grounding, which translates natural language instructions into precise pixel coordinates, is essential for developing practical GUI agents. However, we observe that existing grounding models exhibit significant \textbf{coordinate prediction instability}—minor visual perturbations (e.g., cropping a few pixels) can drastically alter predictions, flipping results between correct and incorrect. This instability severely undermines model performance, especially for samples with high-resolution and small UI elements. To address this issue, we propose Multi-View Prediction (MVP), a training-free framework that enhances grounding performance through multi-view inference. Our key insight is that while single-view predictions may be unstable, aggregating predictions from multiple carefully cropped views can effectively distinguish correct coordinates from outliers. MVP comprises two components: (1) \textbf{Attention-Guided View Proposal}, which derives diverse views guided by instruction-to-image attention scores, and (2) \textbf{Multi-Coordinates Clustering}, which ensembles predictions by selecting the centroid of the densest spatial cluster. Extensive experiments demonstrate MVP's effectiveness across various models and benchmarks. Notably, on ScreenSpot-Pro, MVP boosts UI-TARS-1.5-7B to 56.1\%, GTA1-7B to 61.7\%, Qwen3VL-8B-Instruct to 65.3\%, and Qwen3VL-32B-Instruct to 74.0\%. The code is available at \url{https://github.com/ZJUSCL/MVP}.
\end{abstract}    
\section{Introduction}
The development of automated agents for graphical user interfaces (GUIs) represents a pivotal frontier in artificial general intelligence (AGI) research~\cite{mllmbasedguiagents,brainedgui,gpt4twebagent,guiagents,guiagentssurvey}. These agents fundamentally rely on GUI grounding, which mappings natural language instructions to their corresponding actionable elements within screenshots or live interfaces~\cite{winspot,visualwebbench,sspro,mmbench}.

GUI grounding models are built upon Large Vision Language Models (LVLMs), typically formulating GUI grounding as a generation task, where models output pixel coordinates as text tokens (e.g., “x=123, y=456”)~\cite{seeclick, uitars,cogagent}. However, it is inherently challenging for language models to establish a robust correspondence between visual elements and text coordinates tokens based on instructions~\cite{guigroundingexplicit,guiactor,spatialgui}. Despite extensive training on GUI images through supervised fine-tuning (SFT) or reinforcement learning (RL), grounding models still generate unexplainable erroneous coordinates, particularly when facing high-resolution images and small target UI elements that are difficult to identify.

\begin{figure*}[t]
  \centering
  \begin{subfigure}{0.36\linewidth}
    \includegraphics[width=\linewidth]{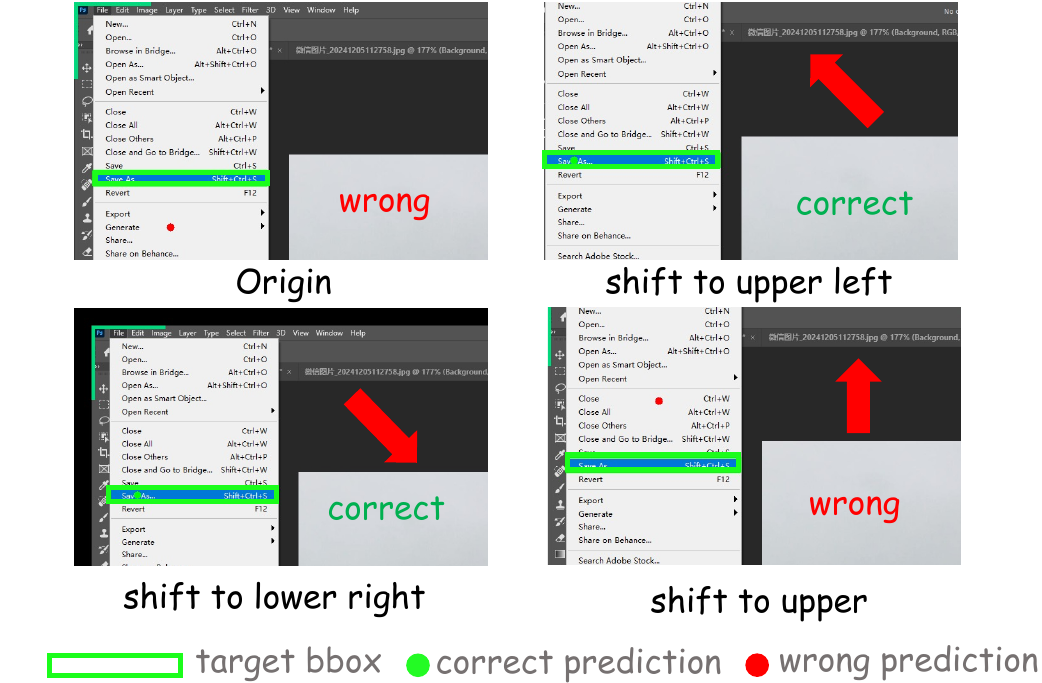}
    \label{fig:instable_example}
    \caption{Example of prediction instability.}
    
  \end{subfigure}
  \hfill
  \begin{subfigure}{0.31\linewidth}
    \includegraphics[width=\linewidth]{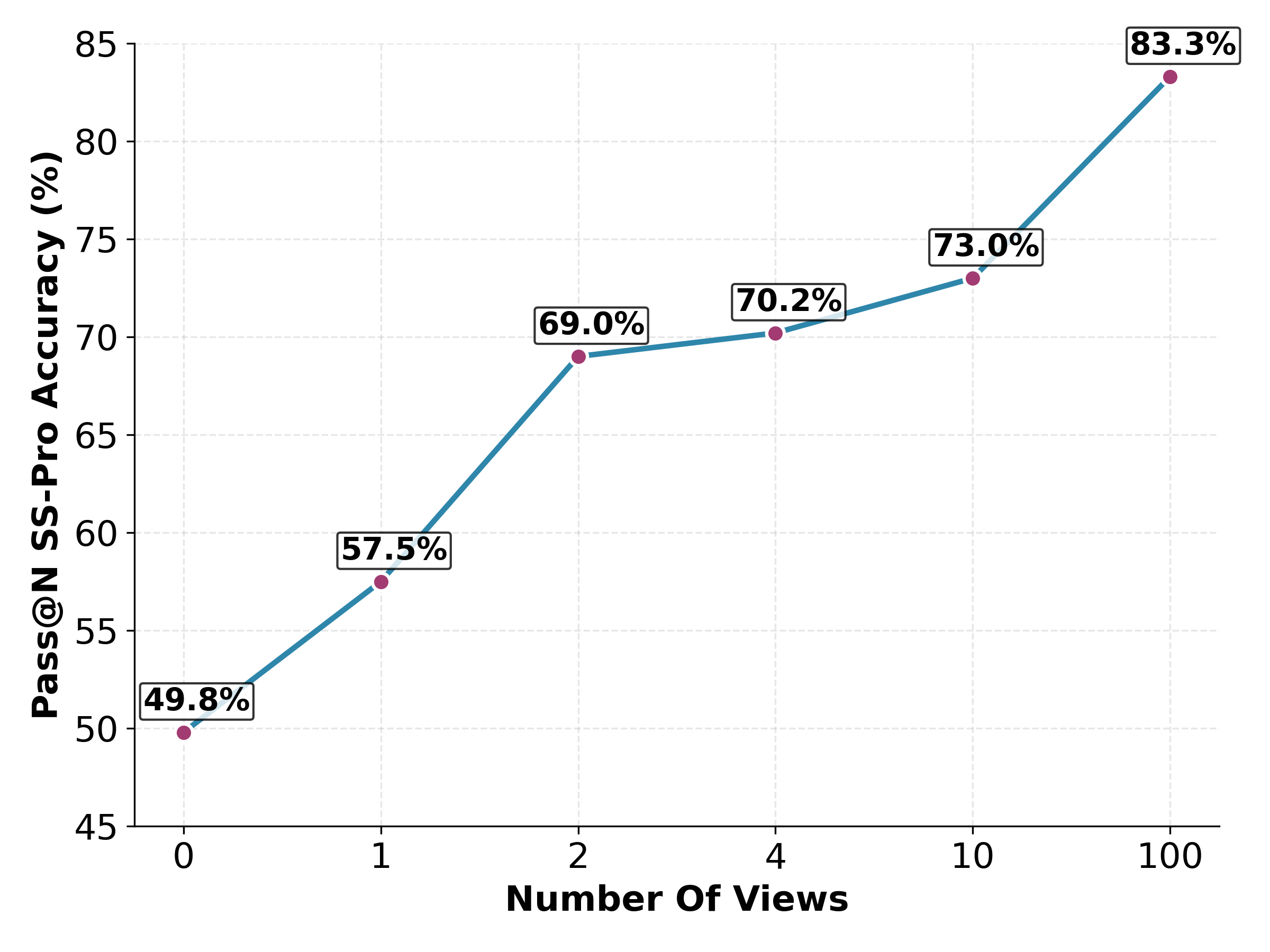}
    \caption{Pass@N increases with number of views.}
    \label{fig:acc_increse}
  \end{subfigure}
  \hfill
  \begin{subfigure}{0.31\linewidth}
    \includegraphics[width=\linewidth]{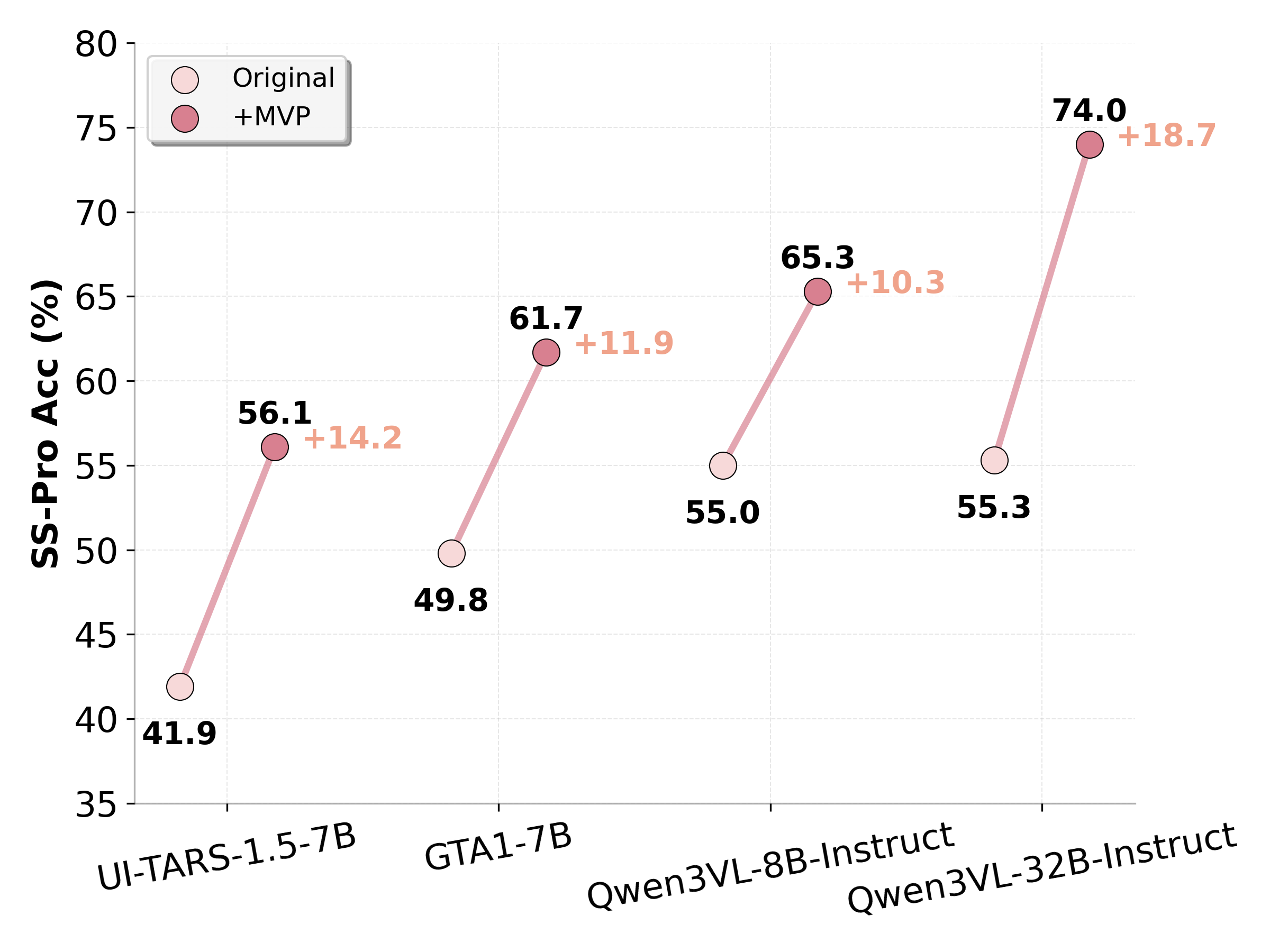}
    \caption{MVP significantly boosts performance.}
    \label{fig:show_rst}
  \end{subfigure}
  
  \caption{(a) An example of model's prediction instability from ScreenSpot-Pro. The instruction is ``save image in a specific format''. Slightly shifting the screenshot causes significantly different predicted coordinates. (b) We crop different views from the original screenshots in ScreenSpot-Pro and then perform inference separately on them using GTA1-7B. The pass@N accuracy improves with number of views increasing, indicting the model possessing the ability to predict the correct prediction. (c) Our MVP significantly improves performance of different architectures and sizes grounding models by aggregating results of different views.}
  \label{fig:analysis}
\end{figure*}

%This paragraph introduce prediction instability
We carefully analyze the failure cases and discover that an incorrect prediction does not mean the model lacks the capability to locate the target. Rather, the models suffer from \textbf{prediction instability}, where minimal perturbations to input images (e.g., shifting by a few pixels) cause dramatic changes in predicted coordinates.
As shown in Figure~\ref{fig:analysis}(a), such minor visual variations can flip predictions between correct and incorrect states, revealing high sensitivity to input perturbations.

%, the model's predicted coordinates change dramatically. This variation may transform an incorrect prediction into a correct one, or conversely, turn a correct prediction into an incorrect one. 

%TODO
%Such instability undermines the model's inherent grounding capability. 

%This paragraph show instability undermines the performance, motivate us to explore multi view prediction.
%This shows that single full-screenshot inference is inadequate to unleash the model's full potential. 
This observation suggests that single full-screenshot inference inadequately unleashes the model's true grounding capability. 
To verify this hypothesis, we conduct a preliminary experiment.
%We further verify this by conducting a preliminary experiment. 
Specifically, we randomly crop multiple 1280×720 sub-regions from the original ScreenSpot-Pro~\cite{sspro} screenshots, ensuring each view contains the target bounding box.
We then predict coordinates for each view. 
As shown in Figure~\ref{fig:analysis}(b), the pass@N accuracy (whether at least one prediction among N views is correct) consistently improves as the number of views increases. This motivate us to leverage multiple sub-regions during inference to improve prediction performance.

%This paragraph introduce MVP pipeline
Based on this observation, we propose the Multiple View Prediction (MVP) framework. It operates in two key stages: Attention-Guided View Proposal and Multi-Coordinate Clustering. First, MVP generates multiple views by cropping sub-regions from the original screenshot, using instruction-to-image attention scores to guide the process. These views maintain diversity while ensuring a high likelihood of containing the target UI elements. Each resulting view, along with the original image, undergoes independent inference to yield multiple coordinate predictions. Finally, the Multi-Coordinate Clustering component aggregates these results by performing spatial clustering on all predicted coordinates and outputs the centroid of the largest cluster as the final prediction.

%This paragraph further explain why multi view can mitigate instability.
The core intuition behind MVP is to mitigate prediction instability through multi-view integration. 
%While we do not know the correctness of each view's predictions, 
Although individual view predictions may be unreliable, they usually exhibit spatial patterns that
the incorrect coordinates tend to scatter arbitrarily whereas the correct ones consistently fall within the target bounding box region. By clustering predictions from diverse views and identifying the densest cluster, MVP effectively distinguishes reliable coordinates from outliers, thereby enhancing grounding performance.

% MVP consists of two key components: the Attention Heuristic Cropping and the Multi-Coordinates Clustering. The former leverages instruction-to-image attention scores from specific layers in the vision-language model, and then selects sub-regions that contain the highest number of top-k score visual tokens as different views. Models inference with these views and generate coordinates separately. The latter performs spatial clustering on the predicted coordinates, selecting the cluster containing the most prediction points and output the center coordinates. 

%This paragraph show performance.
MVP is a training-free framework that can be easily integrated with different grounding models, such as GTA1-7B~\cite{gta1}, UI-TARS-1.5-7B~\cite{ui-tars-15-seed}, and Qwen3VL-\{8B, 32B\}-Instruct~\cite{Qwen3VL}, spanning from 7B/8B to 32B parameter scales. Experimental results on ScreenSpot-Pro~\cite{sspro}, UI-Vision~\cite{uivision} and OS-World-G~\cite{osworldg} benchmarks demonstrate that MVP can significantly improve existing grounding models' performance.

\noindent\textbf{Contributions.} Our contributions are threefold:
\begin{itemize}

\item  We identify coordinate prediction instability in grounding models, which severely undermines model performance.

\item We propose Multi-View Prediction (MVP), a training-free framework that aggregates predictions from multiple attention-guided views through spatial clustering to mitigate prediction instability.

%a training-free framework that aggregates multiple coordinate predictions from different cropped views of the original screenshot to enhance prediction stability and improve performance.

\item We demonstrate that MVP can integrate with grounding models of different architectures, improving accuracy on three challenging grounding benchmarks.
\end{itemize}
\section{Related Work}
GUI grounding, the task of mapping natural language instructions to precise coordinates, is a core capability for developing GUI agents capable of real-world application.

\begin{figure*}[t]
  \centering
  \begin{subfigure}{0.36\linewidth}
    \includegraphics[width=\linewidth]{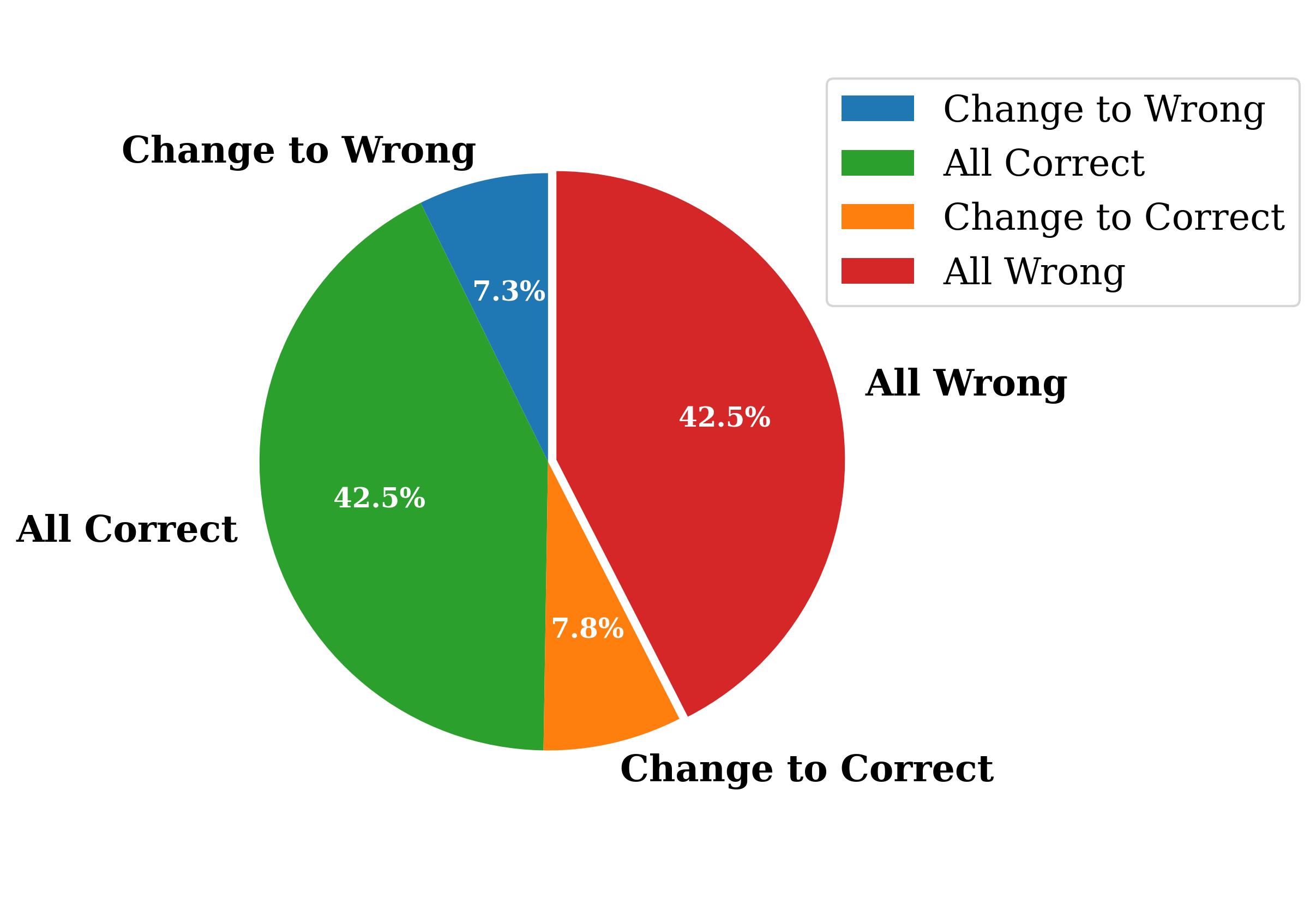}
    \caption{Prediction flips under visual perturbation.}
    \label{fig:analysis_pie_chart}
  \end{subfigure}
  \hfill
  \begin{subfigure}{0.31\linewidth}
    \includegraphics[width=\linewidth]{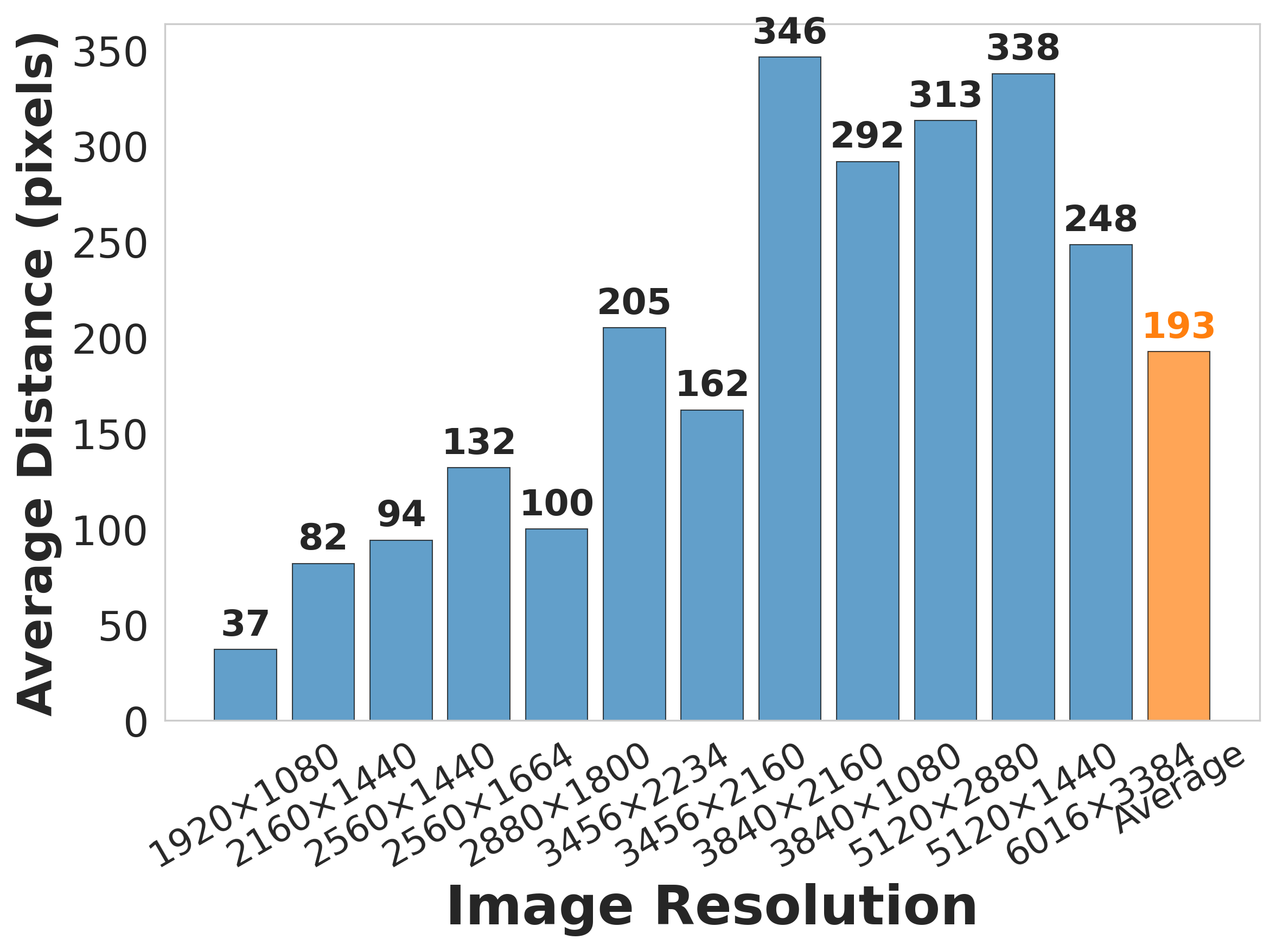}
    \caption{ Instability intensifies with image resolution.}
    \label{fig:analysis_resolution}
  \end{subfigure}
  \hfill
  \begin{subfigure}{0.31\linewidth}
    \includegraphics[width=\linewidth]{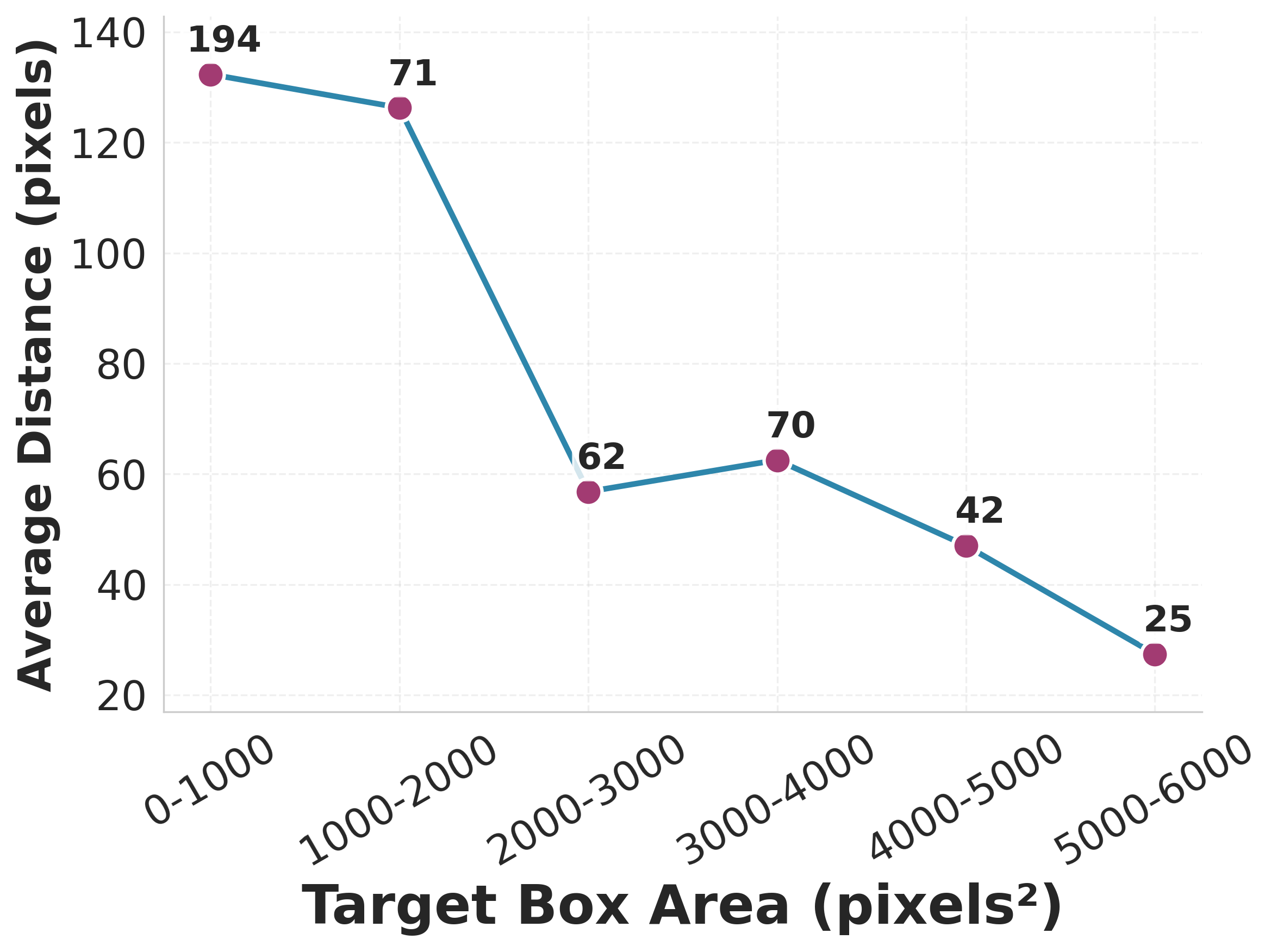}
    \caption{ Instability intensifies with smaller targets.}
    \label{fig:analysis_aera}
  \end{subfigure}
  
  \caption{We evaluate model instability by adding a 28-pixel border to ScreenSpot-Pro images and performing separate inference runs with GTA1-7B. (a) This minor visual perturbation causes 7.3\% of originally correct predictions to become incorrect, and 7.8\% of originally wrong predictions to become correct, revealing high sensitivity to input variations. (b) When analyzing the distance between the two predicted coordinates grouped by image resolution, we observe that instability increases significantly with higher resolutions. (c) Similarly, when grouping by the area of the target region, we find that instability is more pronounced for smaller UI elements.}
  \label{fig:premi}
\end{figure*}

The task was first introduced by SeeClick~\cite{seeclick} through the ScreenSpot benchmark, showing that grounding pretraining improves end-to-end success across mobile, web, and desktop UIs scenarios. Subsequent research in this area can be broadly categorized into following directions:

\begin{itemize}

\item \textbf{Direct Coordinate Optimization} methods enhance the model's grounding ability by supervised fine-tuning (SFT) on large-scale GUI-specific datasets~\cite{ariaui, uground, osatlas, showui, uitars}, or employing reinforcement learning (RL) with rule-based rewards~\cite{gta1, gaussion, guir1}, directly optimizing the output coordinates tokens. While these methods improve performance, they require substantial computational resources for training and still struggle with high resolution and small UI elements. Unlike these resource-intensive approaches, MVP requires no additional training. 

\item \textbf{Iterative Zoom-in} methods reframe grounding as a multi-step decision process. These approaches either leverage execution feedback from GUI agents~\cite{testtime} or exploit the model's own reasoning capability~\cite{spatialgui, guispotlight, uiins, guiarp} to iteratively narrow down to a correct sub-region and then make the final prediction. However, these methods suffer from (1) error accumulation, where a mistake in an early step propagates to subsequent stages; (2) additional training or requiring feedback from external agents. In contrast to this sequential search for one optimal view, our MVP employs a parallel multi-view strategy, aggregating predictions via clustering, thereby avoiding error propagation. Meanwhile, MVP operates in a fully training-free manner without relying on any external feedback.
\item \textbf{Attention-Based} methods leverage the intrinsic instruction–spatial alignment in LVLMs. They derive cross-attention scores from transformer layers to identify the most relevant visual patch and directly output its center coordinates~\cite{guiactor, v2p, attentiondriven}. However, these methods are highly dependent on the precision of the attention scores, limiting their generalizability across different instructions. Different from these methods, MVP preserves the standard text generation paradigm as well as better generalization across diverse scenarios.

\end{itemize}

% Different from Direct Coordinate Optimization methods and simply attention-based methods, our work reveals existing models' sensitivity to minor visual variations, proposing a training-free pipeline that leverages multi-view prediction to more fully exploit a model's inherent grounding potential. In contrast to the iterative zoom-in path that seeks one optimal view, our MVP employs a multi-view strategy, aggregating predictions via clustering to determine the answer.
\section{Preliminary Analysis}

In this section, we systematically diagnose the prediction instability in GUI grounding models. We experiment with GTA1-7B model~\cite{gta1} on the ScreenSpot-Pro benchmark~\cite{sspro} to demonstrate that this instability severely limits model reliability and then discuss on its underlying causes.

\subsection{Single Inference is Unreliable}
Our core finding is that grounding models are highly sensitive to visual perturbations, making single-inference results unreliable. Specifically, by adding a mere 28-pixel border (significantly smaller than the image resolution) to screenshots, we observe drastically different coordinate predictions from the same model: the average coordinate shift of 193 pixels far exceeds the size of typical UI elements in ScreenSpot-Pro (Figure~\ref{fig:premi}(b)).

Crucially, this instability directly impacts accuracy. As shown in Figure~\ref{fig:premi}(a), the model achieved 57.5\% accuracy when considering at least one of the two predictions as correct—significantly higher than its 49.8\% single-prediction accuracy. This gap confirms that the model possesses the requisite capability, but single-view inference fails to harness it consistently.

\subsection{What Drives Instability}
\label{sec:reason}
To understand what makes predictions unstable, we analyze how instability varies with input resolution and target UI element size. Figure \ref{fig:premi}(b) and \ref{fig:premi}(c) reveal a clear trend: instability intensifies dramatically for (1) high-resolution screenshots and (2) samples with small target elements.

We attribute these challenges to both architectural and data-driven limitations. At the architectural level, the task of mapping high-dimensional visual patches to discrete coordinate tokens via a language head is inherently difficult, especially for high-resolution inputs where minor spatial changes yield vastly different token sequences. Additionally, current training datasets lack sufficient examples of high-resolution screenshots and small UI elements samples, creating a generalization shortfall at test time.

This diagnosis naturally leads to our solution: if a single view is unreliable, but the model can sometimes predict correctly, then \textbf{aggregating predictions from multiple views} should yield more robust and accurate results.
\section{Methods}
\begin{figure*}
  \centering
  \begin{subfigure}{0.95\linewidth}
    \includegraphics[width=\linewidth]{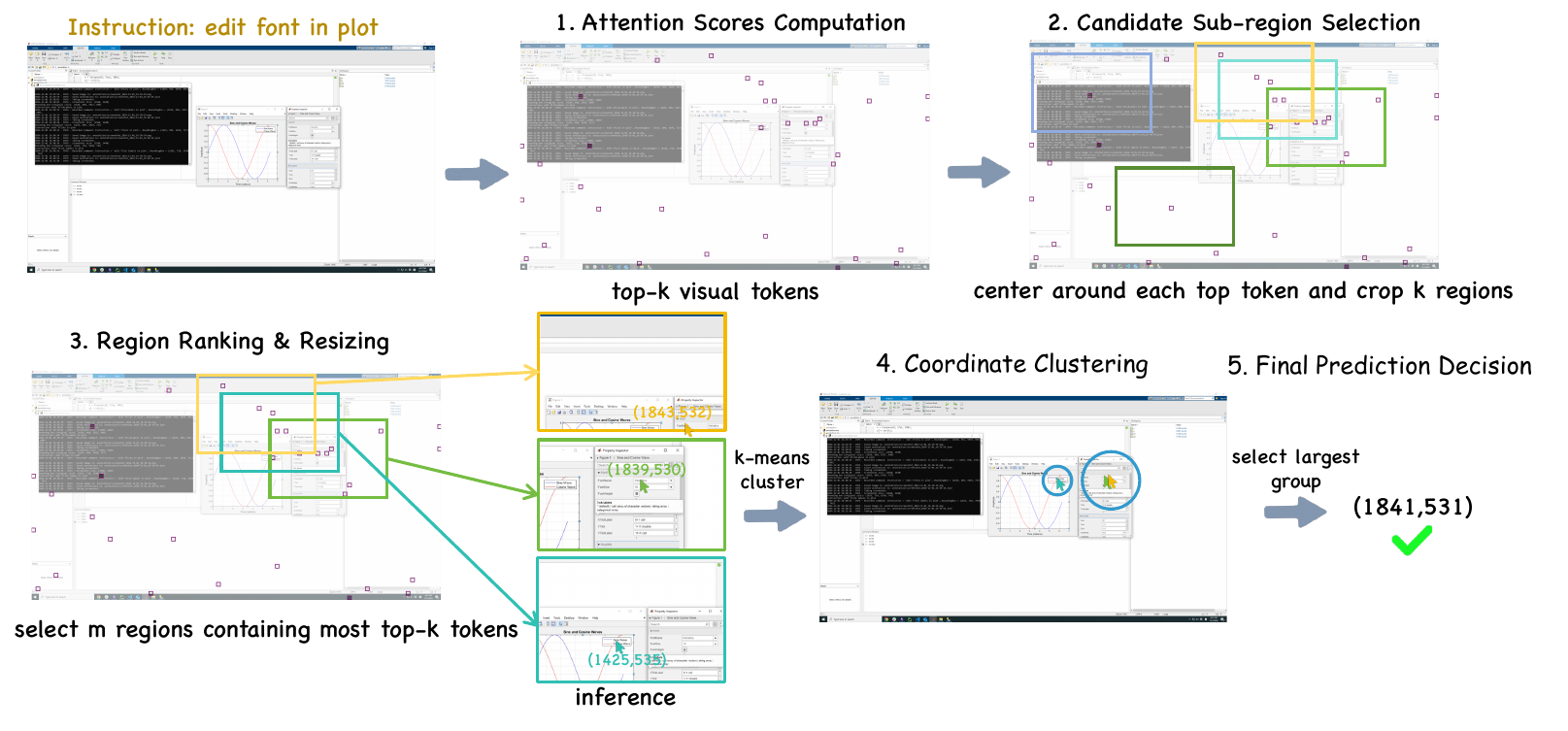}
    \label{fig:instable_example}
  \end{subfigure}
  \vspace{-15pt}
  \caption{Overview of our Multiple View Prediction (MVP) pipeline, which consists of two main stages: Attention-Guided View Proposal and Multi-Coordinate Clustering. First, MVP takes the user instruction and screenshot, forwarding through the language model to derive attention scores from the instruction to each visual token. Then the top-k scores tokens are selected, and an h×w sub-region is cropped around the center of each corresponding visual patch. These sub-regions are ranked by the number of top-k tokens they contain. The top-m regions are chosen and enlarged to form the final set of views. The model independently predicts coordinates for each view. Finally, MVP aggregates all the predictions by clustering the coordinates based on spatial proximity and outputs the center of the largest cluster as the final prediction.}
  \label{fig:archtecture}
\end{figure*}
In this section, we propose the training-free Multiple View Prediction (MVP) framework to address the identified instability issues. The MVP framework consists of two main components: (1) Attention-Guided View Proposal identifies sub-regions containing the target bounding box and generates different views, reducing input resolution and enhancing small target visibility (2) Multi-Coordinate Clustering aggregates the predictions from multiple views and determines the final coordinates, identifying the spatially consistent cluster to filter out outliers and enhance robustness.

\subsection{Attention-Guided View Proposal}

This module aims to generate multiple views by locating sub-regions that contain the target UI elements, guided by cross-attention scores. It leverages the strong text-visual alignment capability of LVLMs, whose attention mechanisms in middle-to-deep layers can effectively localize instruction-relevant regions~\cite{flexselect,prunevid,pyramiddrop,attentiondriven}. Given the system prompt, a GUI screenshot, and the user instruction, this module outputs $m$ cropped image regions. The process consists of three main steps: Attention Score Computation, Candidate Sub-region Selection, and Region Ranking \& Resizing.

\textbf{Attention Score Computation.} We compute the cross-attention scores using the text token as the query and the visual tokens as the keys. Specifically, we use the center comma token (``,'') from the predicted coordinate format (e.g. ``(123,456)'') as the query token, as it demonstrates better region localization performance, with further analysis provided in the Appendix. Let $V \in \mathbb{R}^{H \times L_v \times d}$ represent the visual tokens and $T_{\text{comma}} \in \mathbb{R}^{H \times 1 \times d}$ denote the comma token. The cross-attention scores are computed as:
\begin{equation}
A = \text{Softmax}\left(\frac{T_\text{comma}V^T}{\sqrt{d}}\right), A \in \mathbb{R}^{H \times L_v}
\label{eq:1}
\end{equation}
The final attention score assigned to each visual token is obtained by averaging across all attention heads:
\begin{equation}
\text{scores} = \frac{1}{H}\sum_{i=1}^{H} A[i, :], \quad \text{scores} \in \mathbb{R}^{L_v}
\label{eq:2}
\end{equation}
where $L_v$ is the number of visual tokens, $H$ is the number of attention heads, and $d$ is the dimension of the model.

\textbf{Candidate Sub-region Selection.} We select $k$ candidate regions based on the computed attention scores. Specifically, we choose the top-$k$ visual tokens with the highest scores. Let \( \mathcal{T} = \{ (t_j, x_j, y_j) \mid j=1,\dots,k \} \) denote the set of top-$k$ visual tokens and their corresponding patch center coordinates, for each top-$k$ token, we crop an $h \times w$ sub-region centered at $(x_i, y_i)$ in the original image, resulting in $k$ candidate regions:
\begin{equation}
R_i = \text{Crop}\left(I, x_i - \frac{w}{2}, y_i - \frac{h}{2}, w, h\right), \quad i \in [1,k]
\label{eq:3}
\end{equation}
where $I$ is the original GUI image and $\text{Crop}(I, x, y, w, h)$ function extracts a rectangular region.

\textbf{Region Ranking \& Resizing.} We select $m$ regions from $k$ candidates to form the final views. Regions containing more top-k visual tokens are considered more likely  to contain the target bounding box. We rank the candidate regions by the number of these tokens whose patch center coordinates fall within the region, and choose the top-$m$ regions:
\begin{equation}
\text{rank}(R_i) = \sum_{j=1}^{k} \mathbb{I} \left[ (x_j, y_j) \in R_i \right], i \in [1,k]
\label{eq:4}
\end{equation}

Considering small UI elements poses more instability to grounding models (Section~\ref{sec:reason}), we enlarge the selected regions to enhance the visibility of small targets:
\begin{equation}
R_i^{\text{resized}} = \text{Resize}(R_i, \alpha h, \alpha w), \quad \alpha > 1
\label{eq:5}
\end{equation}

\begin{algorithm}[t]
\caption{Attention-Guided View Proposal}
\label{alg:attention_cropping}

\begin{algorithmic}[1]
\Require Text instruction $T$, original image $I$, view size $(h, w)$, view number $m$, resize ratio $\alpha$
\Ensure Candidate views set $\mathcal{V} = \{R_1^{\text{resized}}, \ldots, R_m^{\text{resized}}\}$

\State \textbf{1. Attention Score Computation}
\State Extract visual tokens $V \in \mathbb{R}^{H \times L_v \times d}$ from $I$
\State Get comma token $T_{\text{comma}} \in \mathbb{R}^{H \times 1 \times d}$ as query token
\State Use Eq~\ref{eq:1}. and Eq~\ref{eq:2}. to compute attention scores
% \State 
% \State \hspace{0.5cm} $A = \text{Softmax}\left(\frac{T_{comma}V^T}{\sqrt{d}}\right)$
% \State \hspace{0.5cm} $scores = \frac{1}{H}\sum_{i=1}^{H}  A[i,:]$

\State \textbf{2. Candidate Sub-region Selection}
\State Sort visual tokens by scores in descending order
\State Select top-$k$ tokens: $\text{tokens} = \{t_1, t_2, \ldots, t_k\}$
\State Get corresponding positions: $\{(x_1, y_1), \ldots, (x_k, y_k)\}$

\State Initialize empty region set $\mathcal{R} = \emptyset$
\For{each selected token $t_i$ at position $(x_i, y_i)$}
    \State  Crop candidate region $\{R_i\}$ using Eq~\ref{eq:3}.
    \State Compute ranking score $rank(R_i)$ using Eq~\ref{eq:4}.
    \State $\mathcal{R} = \mathcal{R} \cup \{R_i\}$
\EndFor

\State \textbf{3. Region Ranking \& Resizing}
\State Sort regions by ranking scores: $R_{(1)}, R_{(2)}, \ldots, R_{(k)}$
\State Select top-$m$ regions: $\mathcal{V} = \{R_{(1)}, \ldots, R_{(m)}\}$
\State Resize each region $R_{(i)}$ in selected set using Eq~\ref{eq:5}.
%     \State $R_{(i)}^{\text{resized}} = \text{Resize}(R_{(i)}, \alpha h, \alpha w)$
% \EndFor

\State \Return $\mathcal{V}$
\end{algorithmic}
\end{algorithm}

\subsection{Multi-Coordinate Clustering}
% The inherent instability of single inference necessitates multi-ensemble aggregating to mitigate it. After obtaining $m$ sub-regions through the cropping module, we perform inference on each sub-region along with the original full image, yielding $m+1$ coordinates $\{(x_i, y_i)\}_{i=1}^{m+1}$. The goal of this module is to determine the final result.
This module takes the $m$ cropped views as input and outputs the final predicted coordinate. It firstly performs inference on each of the $m$ views along with the original full image, yielding $m+1$ coordinate predictions ${(x_i, y_i)}_{i=1}^{m+1}$, then identifies the correct prediction by clustering spatially consistent coordinates and filtering out outliers. This process consists of two steps: Coordinate Clustering and Final Prediction Decision.

\paragraph{Coordinate Clustering} 
% We first group predictions based on spatial proximity using a distance threshold $\tau$. If the distance between two coordinates falls below $\tau$, they are assigned to the same cluster. Formally, for any two predictions $p_i = (x_i, y_i)$ and $p_j = (x_j, y_j)$:
We cluster the coordinate predictions using K-means based on the distance. 
The metric between any two predictions $p_i$ and $p_j$ is calculated as:
\begin{equation}
d(p_i, p_j) = \sqrt{(x_i - x_j)^2 + (y_i - y_j)^2}
\end{equation}

% \begin{equation}
%     d(p_i, p_j) = \sqrt{(x_i - x_j)^2 + (y_i - y_j)^2} < \tau \Rightarrow p_i, p_j \in G_k
% \end{equation}

\paragraph{Final Prediction Decision} The reliability of a prediction cluster $G_k$ is determined by its size $|G_k|$—while incorrect predictions may scatter arbitrarily, correct ones are spatially consistent as they all fall within the target bounding box. We select the center coordinates of the largest cluster as the final prediction:

\begin{equation}
    G^* = \arg\max_{G_k} |G_k|, \quad (x_{\text{final}}, y_{\text{final}}) = \frac{1}{|G^*|}\sum_{p_i \in G^*} p_i
    \label{eq:7}
\end{equation}

In cases when multiple clusters have the same maximum size, we leverage the attention-based ranking for decision, selecting the cluster whose points corresponding regions containing the most top-$k$ visual tokens:

\begin{equation}
G^* = \arg \max_{G_k \in G} \sum{\text{rank}(R_i)}, p_i \in G_k
\label{eq:8}
\end{equation}

\begin{algorithm}[t]
\caption{Multi-Coordinate Clustering}
\label{alg:coordinate_clustering}

\begin{algorithmic}[1]
\Require Distance threshold $\tau$, coordinate set $\mathcal{C} = \{p_1, p_2, \dots, p_{m+1}\}$
\Ensure Final coordinate $p_{\text{final}}$

\State \textbf{Step 1: Cluster coordinates}
\State Initialize clusters $\mathcal{G} = \{\}$, unassigned $U = \mathcal{C}$
\While{$U \neq \emptyset$}
    \State $p_{\text{seed}} \gets U[0]$, $G \gets \{p_{\text{seed}}\}$, $U \gets U \setminus \{p_{\text{seed}}\}$
    \Repeat
        \State $G_{\text{prev}} \gets G$
        \For{$p \in U$}
            \State $\text{center} \gets \frac{1}{|G|} \sum_{q \in G} q$
            \If{$\|p - \text{center}\|_2 \leq \tau$}
                \State $G \gets G \cup \{p\}$, $U \gets U \setminus \{p\}$
            \EndIf
        \EndFor
    \Until{$G = G_{\text{prev}}$}
    \State $\mathcal{G} \gets \mathcal{G} \cup \{G\}$
\EndWhile

\State \textbf{Step 2: Final Prediction Decision}
\State $G^* = \arg\max_{G \in \mathcal{G}} |G|$
\If{$\exists$ multiple $G$ with max size}
    \State $G^* = \arg \max_{G_k \in G} \sum{\text{rank}(R_i)}, p_i \in G_k$
\EndIf
\State $p_{\text{final}} = \frac{1}{|G^*|}\sum_{p_i \in G^*} p_i$
\State \Return $p_{\text{final}}$
\end{algorithmic}
\end{algorithm}

\begin{table*}[t]
\centering
\footnotesize
\setlength{\tabcolsep}{9pt}
\begin{tabular}{@{}lccccccc@{}}
\toprule
\textbf{Model} & \textbf{Development} & \textbf{Creative} & \textbf{CAD} & \textbf{Scientific} & \textbf{Office} & \textbf{OS} & \textbf{Overall} \\
\midrule
\rowcolor{gray!15} \multicolumn{8}{c}{\textbf{Closed-source Models}} \\
GPT-4o~\cite{gpt4o} & 0.7 & 0.6 & 1.5 & 1.2 & 0.9 & 0.0 & 0.8 \\
Claude Computer Use~\cite{claude} & 12.6 & 16.8 & 11.9 & 25.8 & 26.9 & 8.1 & 17.1 \\
UI-TARS-1.5~\cite{ui-tars-15-seed} & 63.9 & 50.4 & 58.2 & 69.3 & 79.6 & 51.0 & 61.6 \\
Seed1.5-VL~\cite{seedvl1.5} & 53.8 & 59.2 & 59.0 & 61.4	 & 74.8 & 60.2 & 60.9 \\
\midrule
\rowcolor{blue!15} \multicolumn{8}{c}{\textbf{Open-Source Models}} \\
SeeClick-7B~\cite{seeclick} & 0.3 & 0.6 & 1.9 & 2.0 & 0.9 & 1.5 & 1.1 \\
UGround-V1-7B~\cite{uground} & 28.1 & 31.7 & 14.6 & 39.0 & 49.6 & 24.5 & 31.1 \\
UGround-V1-72B~\cite{uground} & 31.1 & 35.8 & 13.8 & 50.0 & 51.3 & 25.5 & 34.5 \\
Qwen2.5-VL-32B-Instruct~\cite{qwen25vl} & 48.8 & 42.2 & 31.0 & 55.5 & 64.3 & 50.5 & 48.0 \\
RegionFocus (Qwen2.5VL-72B)~\cite{testtime} & 51.2 & 57.2 & 60.9 & 66.5 & 80.9 & 57.1 & 61.6 \\
% GTA1-32B~\cite{gta1} & 56.2 & 46.3 & 38.7 & 59.1 & 72.2 & 53.1 & 53.6 \\
GTA1-72B~\cite{gta1} & 57.2 & 51.0 & 49.8 & 63.0 & 77.0 & 57.1 & 58.4 \\
GUI-Actor-2.5VL-7B~\cite{guiactor}& 38.1 & 41.3 & 38.3 & 50.8 & 63.0 & 38.8 & 44.6 \\ 
SE-GUI-7B~\cite{segui} & 44.5 & 37.2 & 42.1 & 54.7 & 70.4 & 38.8 & 47.2 \\
UI-Venus-72B~\cite{uivenus} & 59.5 & 55.4 & 57.5 & 66.5 & 77.8 & 57.7 & 61.9 \\
V2P-7B~\cite{v2p} & 46.8 & 43.1 & 47.1 & 56.3 & 68.3 & 45.4 & 50.6 \\ 
GMS (Gemini-2.5-Flash-Lite)~\cite{scanner} & 44.8 & 54.8 & 57.5 & 55.9 & 70.4 & 44.9 & 54.6 \\
GUI-Spotlight~\cite{guispotlight} & 53.3 & 44.4 & 51.0 & 52.4 & 71.3 & 46.9 & 52.8 \\
GUI-Cursor-7B~\cite{guicursor} & 57.5 & 45.8 & 53.2 & 61.4 & 74.8& 50.0& 56.5\\
UI-INS-32B~\cite{uiins} & 55.8 & 46.4 & 48.4 & 62.2& 80.0& 54.1 & 57.0 \\
HyperClick~\cite{hyperclick} & 46.9 & 45.1 &48.5& 56.7& 60.9& 40.8& 48.2 \\
\midrule
UI-TARS-1.5-7B & 36.4 & 38.1 & 20.5 & 49.6 & 68.7 & 31.5 & 41.9 \\
+ MVP & 51.8\textcolor{red}{$\uparrow$15.4} & 50.0\textcolor{red}{$\uparrow$11.9} & 53.3\textcolor{red}{$\uparrow$32.8} & 57.9\textcolor{red}{$\uparrow$8.3} & 73.0\textcolor{red}{$\uparrow$4.3} & 54.6\textcolor{red}{$\uparrow$23.1} & 56.1\textcolor{red}{$\uparrow$14.2} \\
GTA1-7B & 43.4 & 44.8 & 44.4 & 55.9 & 74.8 & 35.2 & 49.8 \\
+ MVP  & 58.9\textcolor{red}{$\uparrow$15.5} & 52.6\textcolor{red}{$\uparrow$7.8} & 60.2\textcolor{red}{$\uparrow$15.8} & 63.0\textcolor{red}{$\uparrow$7.1} & 79.1\textcolor{red}{$\uparrow$4.3} & 56.1\textcolor{red}{$\uparrow$20.9} & 61.7\textcolor{red}{$\uparrow$11.9} \\
Qwen3VL-8B-Instruct & 52.8 & 49.1 & 49.0 & 56.7 & 75.2 & 50.5 & 55.0 \\
+ MVP  & 61.5\textcolor{red}{$\uparrow$8.7} & 60.2\textcolor{red}{$\uparrow$11.1} & 61.3\textcolor{red}{$\uparrow$12.3} & 67.3\textcolor{red}{$\uparrow$10.6} & 82.6\textcolor{red}{$\uparrow$7.4} & 62.8\textcolor{red}{$\uparrow$12.3} & 65.3\textcolor{red}{$\uparrow$10.3} \\
Qwen3VL-32B-Instruct & 43.1 & 54.4 & 57.5 & 62.6 & 73.0 & 42.3 & 55.3 \\
+ MVP  & \textbf{71.6}\textcolor{red}{$\uparrow$28.5} & \textbf{69.3}\textcolor{red}{$\uparrow$14.9} & \textbf{74.7}\textcolor{red}{$\uparrow$17.2} & \textbf{70.5}\textcolor{red}{$\uparrow$7.9} & \textbf{87.4}\textcolor{red}{$\uparrow$14.4} & \textbf{73.5}\textcolor{red}{$\uparrow$31.2} & \textbf{74.0}\textcolor{red}{$\uparrow$18.7} \\
\bottomrule
\end{tabular}
\vspace{4pt}
\caption{Evaluation results on the ScreenSpot-Pro benchmark. Baseline models are evaluated using official instructions, while other models' results are sourced from the benchmark leaderboard. Our method shows significant performance improvements across all categories. These consistent gains across diverse model architectures validate MVP's effectiveness in addressing high-resolution GUI grounding challenges through its multi-view prediction mechanism.}
\label{tab:sspro}
\vspace{-10pt}
\end{table*}
\section{Experiments}
\begin{table*}[t]
\centering
\footnotesize
\setlength{\tabcolsep}{3pt}
\begin{tabular}{@{}lcccccc@{}}
\toprule
\textbf{Model} & \textbf{Text Match} & \textbf{Element Recognition} & \textbf{Layout Understanding} & \textbf{Fine-grained Manipulation} & \textbf{Refusal} & \textbf{Average} \\
\midrule
\rowcolor{gray!15} \multicolumn{7}{c}{\textbf{Closed-source Models}} \\
Operator~\cite{operator} & 51.3 & 42.4 & 46.6 & 31.5 & - & 40.6 \\
Gemini-2.5-pro~\cite{geminipro} & 59.8 & 45.5 & 49.0 & 33.6 & 38.9 & 45.2 \\
Seed1.5-VL~\cite{seedvl1.5} & 73.9 & 66.7 & 69.6 & 47.0 & 18.5 & 62.9 \\
\midrule
\rowcolor{blue!15} \multicolumn{7}{c}{\textbf{Open-Source Models}} \\
Qwen2.5-VL-72B~\cite{qwen25vl} & 52.6 & 74.6 & 74.7 & 55.3 & - & 62.2 \\
% UI-TARS-7B~\cite{uitars} & 20.1 & 24.3 & 8.4 & 17.6\\
UI-TARS-72B~\cite{uitars} & 69.4 & 60.6 & 62.9 & 45.6 & - & 57.1 \\
UI-Venus-Ground-72B~\cite{uivenus} & 82.1 & 71.2 & 70.7 & 64.4 & - & 70.4 \\
OS-Atlas-7B~\cite{osatlas} & 44.1 & 29.4 & 35.2 & 16.8 & 7.4 & 27.7 \\

% \rowcolor{pink!10}
UGround-V1-7B~\cite{uground} & 51.3 & 40.3 & 43.5 & 24.8 & - & 36.4 \\
Aguvis-7B~\cite{aguvis} & 55.9 & 41.2 & 43.9 & 28.2 & - & 38.7 \\
Jedi-7B~\cite{osworldg} & 65.9 & 55.5 & 57.7 & 46.9 & 7.4 & 54.1 \\
GTA1-72B~\cite{gta1} & 57.9 & 76.9 & 77.3 & 66.7 & - & 66.7 \\
GUI-Spotlight~\cite{guispotlight} & 68.2 & 60.6 & 63.2 & 45.6 & - & 62.7 \\
\midrule
UI-TARS-1.5-7B & 67.3 & 64.5 & 65.2 & 42.9 & - & 61.9 \\
+ MVP & 78.5\textcolor{red}{$\uparrow$11.2} & 69.1\textcolor{red}{$\uparrow$4.6} & 70.0\textcolor{red}{$\uparrow$4.8} & 59.7\textcolor{red}{$\uparrow$16.8} & - & 66.8\textcolor{red}{$\uparrow$4.9} \\
GTA1-7B & 80.4 & 69.4 & 69.1 & 60.4 & - & 67.5 \\
+ MVP & 80.8\textcolor{red}{$\uparrow$0.4} & 71.6\textcolor{red}{$\uparrow$2.2} & 71.1\textcolor{red}{$\uparrow$2.0} & 61.0\textcolor{red}{$\uparrow$0.6} & - & 68.7\textcolor{red}{$\uparrow$1.2} \\
Qwen3VL-8B-Instruct & 79.7 & 71.8 & 74.3 & 59.1 & - & 68.8 \\
+ MVP & 82.4\textcolor{red}{$\uparrow$2.7} & 76.7\textcolor{red}{$\uparrow$4.9} & 79.5\textcolor{red}{$\uparrow$5.2} & 61.7\textcolor{red}{$\uparrow$2.6} & - & 72.7\textcolor{red}{$\uparrow$3.9} \\
Qwen3VL-32B-Instruct & 81.6 & 75.7 & 77.1 & 62.4 & - & 71.7 \\
+ MVP & 81.6 & 76.4\textcolor{red}{$\uparrow$0.7} & 77.5\textcolor{red}{$\uparrow$0.4} & 62.4 & - & 72.0\textcolor{red}{$\uparrow$0.3} \\
\bottomrule
\end{tabular}
% \vspace{2pt}
\caption{Evaluation results on the OS-World-G benchmark. The baseline results are evaluated using the corresponding official GitHub instructions, while other models' results are sourced directly from the UI-Venus Technical Report~\cite{uivenus}. These experimental results demonstrate the effectiveness of our MVP framework in real-world interactive scenarios.}
\label{tab:osworld-g}
\vspace{-10pt}
\end{table*}
\subsection{Models and Benchmarks}
We conduct comprehensive evaluations across diverse grounding models and scales to assess the generalizability and effectiveness of our MVP framework, including UI-TARS-1.5-7B~\cite{ui-tars-15-seed}, GTA1-7B~\cite{gta1} and Qwen3VL-\{8B, 32B\}-Instruct~\cite{Qwen3VL}. The models are evaluated on three challenging GUI grounding benchmarks: (1) ScreenSpot-Pro~\cite{sspro}, a benchmark for evaluating visual grounding on high-resolution screenshots of professional software, covering application domains such as creative tools, office. (2) UI-Vision~\cite{uivision}, a diverse benchmark which samples from 83 applications across 6 domains with dense referring expressions platforms. (3) OS-World-G~\cite{osworldg}, a real-world interactive benchmark containing 564 screenshots from OSWorld~\cite{osworld}, focusing on operating-system tasks like file operations and app launching.

\subsection{Implementation Details}
We run 7B/8B model on 8 RTX4090 and 32B model on 4 A100. The view size $(h, w)$ is set to 1280 × 720 for all experiments, and then are resized to 2560 × 1440. The view number $m$ is set to 4 for GTA1-7B and UI-TARS-1.5-7B, and 2 for Qwen3VL-\{8B,32B\}-Instruct by default. For screenshots with resolution lower than 720P, we generate different views by adding 28-pixel black border to the left, right, upper and bottom of the image accordingly, or we apply the Attention Heuristic Cropping to crop views. According to FlexSelect~\cite{flexselect}, we choose attention scores from layer 20 of the language model for GTA1-7B and UI-TARS-1.5-7B, layer 24 for Qwen3VL-8B-Instruct and layer 48 for Qwen3VL-32B-Instruct to crop views. Top-100 visual tokens are selected to crop candidate regions. The K-means threshold $\tau$ is set to 14 pixels for all models. The evaluation prompt follows the model's official GitHub repository.
\subsection{Evaluation Results}
\paragraph{Results on ScreenSpot-Pro} Grounding models exhibit greater instability when processing high-resolution screenshots, leading to suboptimal performance. Our MVP framework effectively addresses this issue through its multi-view prediction mechanism, as demonstrated by consistent performance gains across all categories and model scales on ScreenSpot-Pro (Table~\ref{tab:sspro}). Specifically, MVP improves GTA1-7B by 11.9 points (49.8 → 61.7), UI-TARS-1.5-7B by 14.2 points (41.9 → 56.1), and Qwen3VL-8B-Instruct by 10.3 points (55.0 → 65.3). Most notably, when integrated with Qwen3VL-32B-Instruct, MVP elevates performance from 55.3 to 74.0, establishing new state-of-the-art results that surpass all existing open-source and closed-source models. These substantial improvements across diverse architectures confirm MVP's effectiveness in addressing the challenges of high-resolution GUI grounding through its multi-view prediction mechanism.

\paragraph{Results on OS-World-G} As shown in Table~\ref{tab:osworld-g}, our MVP framework demonstrates consistent performance improvements on OS-World-G, enhancing the average scores of UI-TARS-1.5-7B by 4.9 points (61.9→66.8), GTA1-7B by 1.2 points (67.5→68.7), Qwen3VL-8B-Instruct by 3.9 points (68.8→72.7), and Qwen3VL-32B-Instruct by 0.3 points (71.7→72.0). These results validate MVP's effectiveness in real-world interactive scenarios. The smaller performance gains on OS-World-G relative to UI-Vision and ScreenSpot-Pro stem from its lower-resolution (720P/1080P), which inherently poses less instability to models and consequently limits improvement potential.

\paragraph{Results on UI-Vision} As shown in Table~\ref{tab:ui-vision}, our MVP framework consistently enhances performance across all UI-Vision categories. For 7B models, MVP brings average gains of +3.4 (UI-TARS-1.5-7B) and +4.0 (GTA1-7B) points, with Qwen3VL-8B-Instruct achieving +4.7 points improvement. Most notably, Qwen3VL-32B-Instruct with MVP establishes new state-of-the-art performance with +7.7 points overall (36.4 → 44.1), surpassing even the 72B UI-Venus-Ground-72B (36.8) by +7.3 points. These results demonstrate that our method effectively improves model performance across different applications and platforms, showing strong generalizability across various domains.

\begin{table}[t]
\centering
\footnotesize
\setlength{\tabcolsep}{3pt}
\begin{tabular}{@{}lcccc@{}}
\toprule
\textbf{Model} & \textbf{Basic} & \textbf{Functional} & \textbf{Spatial} & \textbf{Average} \\
\midrule
\rowcolor{gray!15} \multicolumn{5}{c}{\textbf{Closed-source Models}} \\
GPT-4o~\cite{gpt4o} & 1.6 & 1.5 & 1.0 & 1.4 \\
Claude-3.7-Sonnet~\cite{Claude3.7} & 9.5 & 7.7 & 7.6 & 8.3 \\
\midrule
\rowcolor{blue!15} \multicolumn{5}{c}{\textbf{Open-Source Models}} \\
Qwen2.5-VL-7B~\cite{qwen25vl} & 1.2 & 0.8 & 0.5 & 0.9 \\ 
UI-TARS-7B~\cite{uitars} & 20.1 & 24.3 & 8.4 & 17.6 \\
UI-TARS-72B~\cite{uitars} & 31.4 & 30.5 & 14.7 & 25.5 \\
UI-Venus-Ground-7B~\cite{uivenus} & 36.1 & 32.8 & 11.9 & 26.5 \\
UI-Venus-Ground-72B~\cite{uivenus} & 45.6 & 42.3 & 23.7 & 36.8 \\
OS-Atlas-7B~\cite{osatlas} & 12.2 & 11.2 & 3.7 & 9.0 \\
Phi-Ground-7B-16C-DPO~\cite{phiground} & 36.8 & 37.1 & 7.6 & 27.2 \\
UGround-V1-7B~\cite{uground} & 15.4 & 17.1 & 6.3 & 12.9 \\
HyperClick~\cite{hyperclick} & 35.3 & 32.1 & 11.0 & 25.7 \\
GUI-Spotlight~\cite{guispotlight} & 32.1 & 30.2 & 9.1 & 23.4 \\
\midrule
UI-TARS-1.5-7B & 29.3 & 27.4 & 10.0 & 22.2 \\
+ MVP & 31.8\textcolor{red}{$\uparrow$2.5} & 31.8\textcolor{red}{$\uparrow$4.4} & 13.2\textcolor{red}{$\uparrow$3.2} & 25.6\textcolor{red}{$\uparrow$3.4} \\
GTA1-7B & 35.2 & 32.9 & 11.5 & 26.5 \\
+ MVP & 38.9\textcolor{red}{$\uparrow$3.7} & 37.7\textcolor{red}{$\uparrow$4.8} & 14.9\textcolor{red}{$\uparrow$3.4} & 30.5\textcolor{red}{$\uparrow$4.0} \\
Qwen3VL-8B-Instruct & 32.9 & 34.1 & 14.7 & 27.2 \\
+ MVP & 38.3\textcolor{red}{$\uparrow$5.4} & 38.8\textcolor{red}{$\uparrow$4.7} & 18.7\textcolor{red}{$\uparrow$4.0} & 31.9\textcolor{red}{$\uparrow$4.7} \\
Qwen3VL-32B-Instruct & 43.9 & 42.8 & 22.4 & 36.4 \\
+ MVP & \textbf{49.4}\textcolor{red}{$\uparrow$5.5} & \textbf{52.0}\textcolor{red}{$\uparrow$9.2} & \textbf{30.8}\textcolor{red}{$\uparrow$8.4} & \textbf{44.1}\textcolor{red}{$\uparrow$7.7} \\
\bottomrule
\end{tabular}
%\vspace{6pt}
\caption{Evaluation results on UI-Vision. Baseline models are evaluated using the official GitHub instructions, while results for other models are taken directly from the UI-Vision leaderboard~\cite{uivision}. Our method effectively improves model performance across diverse applications, exhibiting strong generalizability.}
\label{tab:ui-vision}
\end{table}

\subsection{Ablations}
\paragraph{Influence of Attention-Guided View Proposal}
This study ablates the Attention-Guided View Proposal module, which generates diverse views by preserving semantically rich regions at a reduced input resolution. We compare it against a Border Padding strategy that creates views by adding 28-pixel borders around the original screenshot. Results in Table~\ref{tab:compare_AHC} show that while Border Padding improves upon the baseline, our Attention-Guided View Proposal method delivers a significantly larger gain. This demonstrates that our attention-guided strategy for selecting informative sub-regions is effective on view generation.

\begin{table}[t]
\centering
\footnotesize
\begin{tabular}{ccc}
\toprule
\textbf{View Proposal Method} & \textbf{Number of Views} & \textbf{SS-Pro Avg.} \\
\midrule
Baseline (Single Full Image) & -- & 49.8 \\
Border Padding & 4 & 57.3 \\
Attention-Guided View Proposal & 4 & \textbf{61.7} \\

\bottomrule
\end{tabular}
\caption{Comparison of view generation methods on the ScreenSpot-Pro benchmark using GTA1-7B. Our Attention Heuristic Cropping method significantly outperforms both the Border Padding strategy and the single full image baseline, underscoring the importance of content-aware cropping over simple spatial augmentation.}
\vspace{-10pt}
\label{tab:compare_AHC}
\end{table}

\paragraph{Influence of Multi-Coordinate Clustering}
This ablation study evaluates the efficacy of our Multi-Coordinate Clustering module in mitigating prediction instability by aggregating coordinates from multiple views. We compare it against two alternative strategies: (1) randomly selecting one prediction, and (2) averaging over all predicted coordinates. As presented in Table~\ref{tab:compare_aggragation}, our clustering-based method achieves a SS-Pro Avg. of 61.7, substantially outperforming both alternatives and the baseline. This result confirms that identifying spatial consensus through clustering is a more effective aggregation strategy than simple averaging or random selection for determining the final prediction, since incorrect predictions may scatter arbitrarily, correct ones are spatially consistent as they all fall within the target bounding box.

\begin{table}[t]
\centering
\footnotesize
\begin{tabular}{ccc}
\toprule
\textbf{Aggregation Method} & \textbf{View Number} & \textbf{SS-Pro Avg.} \\
\midrule
Baseline (Single Full Image) & -- & 49.8 \\
Average of Coordinates & 4 & 46.6 \\
Random Selection & 4 & 55.7 \\
Multi-Coordinate Clustering & 4 & \textbf{61.7} \\
\bottomrule
\end{tabular}
\caption{Comparison of aggregation strategies on the ScreenSpot-Pro benchmark using GTA1-7B. Our proposed Multi-Coordinates Clustering method demonstrates superior performance, highlighting the importance of robust spatial consensus over naive averaging or random selection.}
\vspace{-10pt}
\label{tab:compare_aggragation}
\end{table}

\paragraph{Influence of View Resizing}
This experiment evaluates the impact of resizing cropped views to a higher resolution (2× of the original size) to enhance the visibility of small target UI elements, thereby mitigating model instability on small regions. We compare the performance with and without this resizing operation. As shown in Table~\ref{tab:compare_resize}, resizing improves the SS-Pro Avg. score by 2.6 points (from 59.1 to 61.7), confirming its contribution to final performance.

\begin{table}[t]
\centering
\footnotesize
\begin{tabular}{ccc}
\toprule
\textbf{Method} & \textbf{Number of Views} & \textbf{SS-Pro Avg.} \\
\midrule
Baseline (Single Full Image) & -- & 49.8 \\
Without Resizing & 4 & 59.1 \\
With Resizing & 4 & \textbf{61.7} \\
\bottomrule
\end{tabular}
\caption{Ablation study on view resizing using the ScreenSpot-Pro benchmark and GTA1-7B. Enlarging the cropped views to twice their original size improves the visibility of target UI elements and yields a performance gain of 2.6 points.}
\label{tab:compare_resize}
\vspace{-12pt}
\end{table}

\begin{figure}[t]
    \centering
    \includegraphics[width=0.9\linewidth]{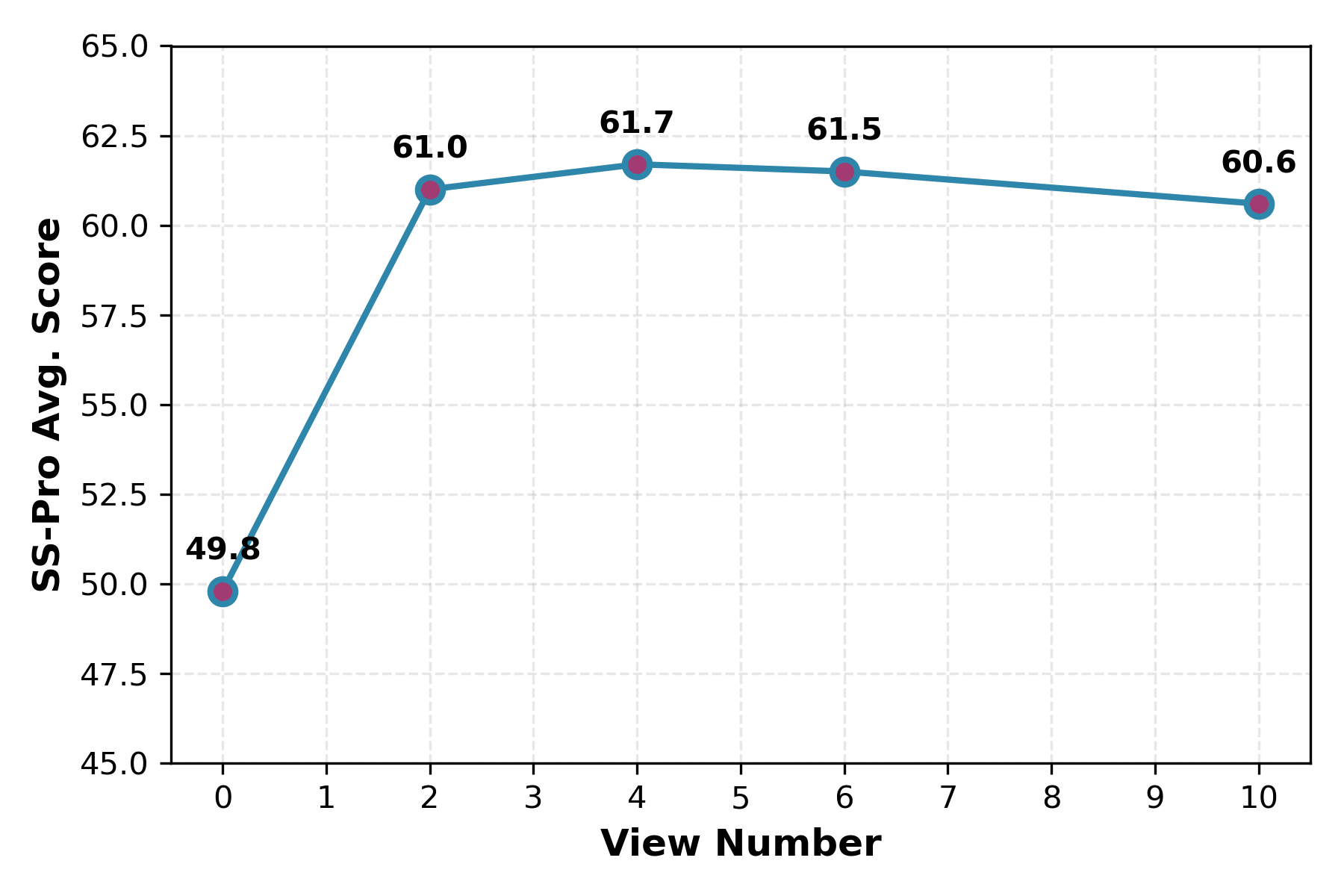}
    \vspace{-10pt}
    \caption{We evaluate GTA1-7B on ScreenSpot-Pro under different view number. Increasing the number of views does not consistently lead to performance improvements. We suggest 4 views as the optimal configuration}
    \label{fig:view_number_effect}
    \vspace{-10pt}
\end{figure}

\paragraph{Influence of View Number}
We compare the performance under different view numbers. As shown in Figure~\ref{fig:view_number_effect}, we observe that increasing the number of views does not consistently lead to performance improvements. Although different views generate distinct coordinates, they tend to cluster around several fixed positions. Consequently, adding more views has limited impact on the final clustering results while increases the time cost. We suggest 4 views as the optimal configuration.

% \begin{table}[H]
% \centering
% \footnotesize
% \setlength{\tabcolsep}{16pt}
% \begin{tabular}{ccc}
% \toprule
% \textbf{Method} & \textbf{View Number} & \textbf{SS-Pro Avg.} \\
% \midrule
% MVP & 3 & 61.0 \\
% MVP & 5 & 61.7  \\
% MVP & 7 & 61.5  \\
% MVP & 11 & 60.6  \\
% Origin & 1 & 49.8  
% \end{tabular}
% \caption{Comparison when using different view number.}
% \label{tab:view_number}
% \end{table}

\section{Conclusion}
In this paper, We find that current existing grounding models are highly sensitive to visual input variations. Even slight perturbations may flip the result between correct and incorrect, which indicates significant prediction instability and undermines model performance. To address this, we propose the training-free Multiple View Prediction (MVP) framework. MVP stabilizes predictions by cropping multiple views of screenshots, obtaining independent predictions, and aggregating them via spatial clustering. Extensive experiments on ScreenSpot-Pro, UI-Vision, and OS-World-G validate that MVP consistently and significantly boosts the performance of diverse grounding models, effectively unleashing their potential without any retraining.

% This paper identifies a critical limitation in GUI grounding models: significant prediction instability, where minor visual perturbations can flip results between correct and incorrect. To address this, we propose the training-free Multi-View Prediction (MVP) framework. MVP stabilizes predictions by generating multiple views of a screenshot, obtaining independent coordinate predictions, and aggregating them via spatial clustering. Extensive experiments on ScreenSpot-Pro, UI-Vision, and OS-World-G benchmarks validate that MVP consistently and significantly boosts the performance of diverse grounding models, effectively unleashing their latent potential without any retraining.

{
    \small
    \bibliographystyle{ieeenat_fullname}
    \bibliography{main}

\begin{thebibliography}{51}
\providecommand{\natexlab}[1]{#1}
\providecommand{\url}[1]{\texttt{#1}}
\expandafter\ifx\csname urlstyle\endcsname\relax
  \providecommand{\doi}[1]{doi: #1}\else
  \providecommand{\doi}{doi: \begingroup \urlstyle{rm}\Url}\fi

\bibitem[{anthropic}(2024)]{claude}
{anthropic}.
\newblock Introducing computer use, a new claude 3.5 sonnet, and claude 3.5 haiku, 2024.
\newblock Accessed: 2024-10-22.

\bibitem[{Anthropic}(2025)]{Claude3.7}
{Anthropic}.
\newblock Claude 3.7 sonnet and claude code, 2025.
\newblock Accessed: 2025-02-25.

\bibitem[Bai et~al.(2025)Bai, Chen, Liu, Wang, Ge, Song, Dang, Wang, Wang, Tang, Zhong, Zhu, Yang, Li, Wan, Wang, Ding, Fu, Xu, Ye, Zhang, Xie, Cheng, Zhang, Yang, Xu, and Lin]{qwen25vl}
Shuai Bai, Keqin Chen, Xuejing Liu, Jialin Wang, Wenbin Ge, Sibo Song, Kai Dang, Peng Wang, Shijie Wang, Jun Tang, Humen Zhong, Yuanzhi Zhu, Mingkun Yang, Zhaohai Li, Jianqiang Wan, Pengfei Wang, Wei Ding, Zheren Fu, Yiheng Xu, Jiabo Ye, Xi Zhang, Tianbao Xie, Zesen Cheng, Hang Zhang, Zhibo Yang, Haiyang Xu, and Junyang Lin.
\newblock Qwen2.5-vl technical report, 2025.

\bibitem[Chen et~al.(2025{\natexlab{a}})Chen, Chen, Wang, Gan, Zhuang, and Gu]{v2p}
Jikai Chen, Long Chen, Dong Wang, Leilei Gan, Chenyi Zhuang, and Jinjie Gu.
\newblock V2p: From background suppression to center peaking for robust gui grounding task, 2025{\natexlab{a}}.

\bibitem[Chen et~al.(2025{\natexlab{b}})Chen, Zhou, Cai, Zhang, Tong, Kong, Zhang, Liu, Liu, Wang, Wang, Jin, and Hoi]{uiins}
Liangyu Chen, Hanzhang Zhou, Chenglin Cai, Jianan Zhang, Panrong Tong, Quyu Kong, Xu Zhang, Chen Liu, Yuqi Liu, Wenxuan Wang, Yue Wang, Qin Jin, and Steven Hoi.
\newblock Ui-ins: Enhancing gui grounding with multi-perspective instruction-as-reasoning, 2025{\natexlab{b}}.

\bibitem[Cheng et~al.(2024)Cheng, Sun, Chu, Xu, YanTao, Zhang, and Wu]{seeclick}
Kanzhi Cheng, Qiushi Sun, Yougang Chu, Fangzhi Xu, Li YanTao, Jianbing Zhang, and Zhiyong Wu.
\newblock {S}ee{C}lick: Harnessing {GUI} grounding for advanced visual {GUI} agents.
\newblock In \emph{Proceedings of the 62nd Annual Meeting of the Association for Computational Linguistics (Volume 1: Long Papers)}, pages 9313--9332, Bangkok, Thailand, 2024. Association for Computational Linguistics.

\bibitem[Gou et~al.(2025)Gou, Wang, Zheng, Xie, Chang, Shu, Sun, and Su]{uground}
Boyu Gou, Ruohan Wang, Boyuan Zheng, Yanan Xie, Cheng Chang, Yiheng Shu, Huan Sun, and Yu Su.
\newblock Navigating the digital world as humans do: Universal visual grounding for {GUI} agents.
\newblock In \emph{The Thirteenth International Conference on Learning Representations}, 2025.

\bibitem[Gu et~al.(2025)Gu, Zeng, Xu, Zhou, Shen, Liu, Zhou, Meng, Xia, Chen, Wen, Dou, Tang, Lin, Liu, Guo, Gong, Jia, Gao, Guo, Deng, Guo, Chen, and Wang]{uivenus}
Zhangxuan Gu, Zhengwen Zeng, Zhenyu Xu, Xingran Zhou, Shuheng Shen, Yunfei Liu, Beitong Zhou, Changhua Meng, Tianyu Xia, Weizhi Chen, Yue Wen, Jingya Dou, Fei Tang, Jinzhen Lin, Yulin Liu, Zhenlin Guo, Yichen Gong, Heng Jia, Changlong Gao, Yuan Guo, Yong Deng, Zhenyu Guo, Liang Chen, and Weiqiang Wang.
\newblock Ui-venus technical report: Building high-performance ui agents with rft, 2025.

\bibitem[Guo et~al.(2025)Guo, Wu, and Feida~Zhu]{seedvl1.5}
Dong Guo, Faming Wu, and etc. Feida~Zhu.
\newblock Seed1.5-vl technical report, 2025.

\bibitem[Hong et~al.(2024)Hong, Wang, Lv, Xu, Yu, Ji, Wang, Wang, Zhang, Li, Xu, Dong, Ding, and Tang]{cogagent}
Wenyi Hong, Weihan Wang, Qingsong Lv, Jiazheng Xu, Wenmeng Yu, Junhui Ji, Yan Wang, Zihan Wang, Yuxuan Zhang, Juanzi Li, Bin Xu, Yuxiao Dong, Ming Ding, and Jie Tang.
\newblock Cogagent: A visual language model for gui agents.
\newblock In \emph{Conference on Computer Vision and Pattern Recognition}, 2024.

\bibitem[Huang et~al.(2025)Huang, Zhou, and Han]{prunevid}
Xiaohu Huang, Hao Zhou, and Kai Han.
\newblock Prunevid: Visual token pruning for efficient video large language models.
\newblock In \emph{Annual Meeting of the Association for Computational Linguistics}, 2025.

\bibitem[Hui et~al.(2025)Hui, Li, Zhao, Banbury, Chen, and Koishida]{winspot}
Zheng Hui, Yinheng Li, Dan Zhao, Colby Banbury, Tianyi Chen, and Kazuhito Koishida.
\newblock {W}in{S}pot: {GUI} grounding benchmark with multimodal large language models.
\newblock In \emph{Proceedings of the 63rd Annual Meeting of the Association for Computational Linguistics (Volume 2: Short Papers)}, pages 1086--1096, Vienna, Austria, 2025. Association for Computational Linguistics.

\bibitem[Lei et~al.(2025)Lei, Xu, Payani, Hong, Liao, Cao, and Ding]{guispotlight}
Bin Lei, Nuo Xu, Ali Payani, Mingyi Hong, Chunhua Liao, Yu Cao, and Caiwen Ding.
\newblock \textsc{GUI-Spotlight}: Adaptive iterative focus refinement for enhanced gui visual grounding, 2025.

\bibitem[Li et~al.(2025{\natexlab{a}})Li, Ziyang, Lin, Luo, Tian, Ma, Huang, and Chua]{sspro}
Kaixin Li, Meng Ziyang, Hongzhan Lin, Ziyang Luo, Yuchen Tian, Jing Ma, Zhiyong Huang, and Tat-Seng Chua.
\newblock Screenspot-pro: {GUI} grounding for professional high-resolution computer use.
\newblock In \emph{Workshop on Reasoning and Planning for Large Language Models}, 2025{\natexlab{a}}.

\bibitem[Li et~al.(2025{\natexlab{b}})Li, Song, Wang, Xiong, Yuan, and Cai]{scanner}
Zhecheng Li, Guoxian Song, Yiwei Wang, Zhen Xiong, Junsong Yuan, and Yujun Cai.
\newblock Generalist scanner meets specialist locator: A synergistic coarse-to-fine framework for robust gui grounding, 2025{\natexlab{b}}.

\bibitem[Lin et~al.(2025)Lin, Li, Gao, Yang, Wu, Bai, Lei, Wang, and Shou]{showui}
Kevin~Qinghong Lin, Linjie Li, Difei Gao, Zhengyuan Yang, Shiwei Wu, Zechen Bai, Weixian Lei, Lijuan Wang, and Mike~Zheng Shou.
\newblock Showui: One vision-language-action model for gui visual agent.
\newblock In \emph{Conference on Computer Vision and Pattern Recognition}, 2025.

\bibitem[Liu et~al.(2024)Liu, Song, Lin, Lam, Neubig, Li, and Yue]{visualwebbench}
Junpeng Liu, Yifan Song, Bill~Yuchen Lin, Wai Lam, Graham Neubig, Yuanzhi Li, and Xiang Yue.
\newblock Visualwebbench: How far have multimodal llms evolved in web page understanding and grounding?, 2024.

\bibitem[Luo et~al.(2025{\natexlab{a}})Luo, Wang, He, Chen, Li, and Xia]{guir1}
Run Luo, Lu Wang, Wanwei He, Longze Chen, Jiaming Li, and Xiaobo Xia.
\newblock Gui-r1 : A generalist r1-style vision-language action model for gui agents, 2025{\natexlab{a}}.

\bibitem[Luo et~al.(2025{\natexlab{b}})Luo, Logeswaran, Johnson, and Lee]{testtime}
Tiange Luo, Lajanugen Logeswaran, Justin Johnson, and Honglak Lee.
\newblock Visual test-time scaling for gui agent grounding, 2025{\natexlab{b}}.

\bibitem[Nayak et~al.(2025)Nayak, Jian, Lin, Rodriguez, Kalsi, Awal, Chapados, Özsu, Agrawal, Vazquez, Pal, Taslakian, Gella, and Rajeswar]{uivision}
Shravan Nayak, Xiangru Jian, Kevin~Qinghong Lin, Juan~A. Rodriguez, Montek Kalsi, Rabiul Awal, Nicolas Chapados, M.~Tamer Özsu, Aishwarya Agrawal, David Vazquez, Christopher Pal, Perouz Taslakian, Spandana Gella, and Sai Rajeswar.
\newblock Ui-vision: A desktop-centric gui benchmark for visual perception and interaction.
\newblock In \emph{Conference on International Conference on Machine Learning}, 2025.

\bibitem[Nguyen et~al.(2025)Nguyen, Chen, Wang, Wu, Park, Hu, Lyu, Wu, Aponte, Xia, Li, Shi, Chen, Lai, Xie, Kim, Zhang, Yu, Tanjim, Ahmed, Mathur, Yoon, Yao, Kveton, Kil, Nguyen, Bui, Zhou, Rossi, and Dernoncourt]{guiagentssurvey}
Dang Nguyen, Jian Chen, Yu Wang, Gang Wu, Namyong Park, Zhengmian Hu, Hanjia Lyu, Junda Wu, Ryan Aponte, Yu Xia, Xintong Li, Jing Shi, Hongjie Chen, Viet~Dac Lai, Zhouhang Xie, Sungchul Kim, Ruiyi Zhang, Tong Yu, Mehrab Tanjim, Nesreen~K. Ahmed, Puneet Mathur, Seunghyun Yoon, Lina Yao, Branislav Kveton, Jihyung Kil, Thien~Huu Nguyen, Trung Bui, Tianyi Zhou, Ryan~A. Rossi, and Franck Dernoncourt.
\newblock Gui agents: A survey.
\newblock In \emph{Proceedings of Annual Meeting of the Association for Computational Linguistics}, 2025.

\bibitem[{OpenAI}(2024)]{gpt4o}
{OpenAI}.
\newblock Hello {GPT}-4o, 2024.
\newblock Accessed: 2024-05-20.

\bibitem[{OpenAI}(2025)]{operator}
{OpenAI}.
\newblock Introducing operator, 2025.
\newblock Accessed: 2025-01-23.

\bibitem[Qin et~al.(2025)Qin, Ye, Fang, Wang, Liang, Tian, Zhang, Li, Li, Huang, Zhong, Li, Yang, Miao, Lin, Liu, Jiang, Ma, Li, Xiao, Cai, Li, Zheng, Jin, Li, Zhou, Wang, Chen, Li, Yang, Liu, Lin, Peng, Liu, and Shi]{uitars}
Yujia Qin, Yining Ye, Junjie Fang, Haoming Wang, Shihao Liang, Shizuo Tian, Junda Zhang, Jiahao Li, Yunxin Li, Shijue Huang, Wanjun Zhong, Kuanye Li, Jiale Yang, Yu Miao, Woyu Lin, Longxiang Liu, Xu Jiang, Qianli Ma, Jingyu Li, Xiaojun Xiao, Kai Cai, Chuang Li, Yaowei Zheng, Chaolin Jin, Chen Li, Xiao Zhou, Minchao Wang, Haoli Chen, Zhaojian Li, Haihua Yang, Haifeng Liu, Feng Lin, Tao Peng, Xin Liu, and Guang Shi.
\newblock Ui-tars: Pioneering automated gui interaction with native agents, 2025.

\bibitem[{qwen}(2025)]{Qwen3VL}
{qwen}.
\newblock Qwen3vl, 2025.

\bibitem[Seed(2025)]{ui-tars-15-seed}
ByteDance Seed.
\newblock Ui-tars-1.5.
\newblock \url{https://seed-tars.com/1.5}, 2025.

\bibitem[Tang et~al.(2025{\natexlab{a}})Tang, Gu, Lu, Liu, Shen, Meng, Wang, Zhang, Shen, Lu, Xiao, and Zhuang]{gaussion}
Fei Tang, Zhangxuan Gu, Zhengxi Lu, Xuyang Liu, Shuheng Shen, Changhua Meng, Wen Wang, Wenqi Zhang, Yongliang Shen, Weiming Lu, Jun Xiao, and Yueting Zhuang.
\newblock Gui-g$^2$: Gaussian reward modeling for gui grounding, 2025{\natexlab{a}}.

\bibitem[Tang et~al.(2025{\natexlab{b}})Tang, Xu, Zhang, Chen, Wu, Shen, Zhang, Hou, Tan, Yan, Song, Shao, Lu, Xiao, and Zhuang]{mllmbasedguiagents}
Fei Tang, Haolei Xu, Hang Zhang, Siqi Chen, Xingyu Wu, Yongliang Shen, Wenqi Zhang, Guiyang Hou, Zeqi Tan, Yuchen Yan, Kaitao Song, Jian Shao, Weiming Lu, Jun Xiao, and Yueting Zhuang.
\newblock A survey on (m)llm-based gui agents, 2025{\natexlab{b}}.

\bibitem[Tao et~al.(2025)Tao, Qin, You, Sui, and Wang]{dycoke}
Keda Tao, Can Qin, Haoxuan You, Yang Sui, and Huan Wang.
\newblock Dycoke: Dynamic compression of tokens for fast video large language models.
\newblock In \emph{Conference on Computer Vision and Pattern Recognition}, pages 18992--19001, 2025.

\bibitem[Team et~al.(2024)Team, Anil, Borgeaud, Alayrac, Yu, Soricut, Schalkwyk, Dai, Hauth, and et~al]{geminipro}
Gemini Team, Rohan Anil, Sebastian Borgeaud, Jean-Baptiste Alayrac, Jiahui Yu, Radu Soricut, Johan Schalkwyk, Andrew~M. Dai, Anja Hauth, and et al.
\newblock Gemini: A family of highly capable multimodal models, 2024.

\bibitem[Wang et~al.(2025{\natexlab{a}})Wang, Liu, Chen, Zhou, Gan, Zeng, Che, Yu, Hao, Shao, Wang, Wu, Wang, Tang, and Hao]{guiagents}
Shuai Wang, Weiwen Liu, Jingxuan Chen, Yuqi Zhou, Weinan Gan, Xingshan Zeng, Yuhan Che, Shuai Yu, Xinlong Hao, Kun Shao, Bin Wang, Chuhan Wu, Yasheng Wang, Ruiming Tang, and Jianye Hao.
\newblock Gui agents with foundation models: A comprehensive survey, 2025{\natexlab{a}}.

\bibitem[Wang et~al.(2025{\natexlab{b}})Wang, Zhang, Masry, Pal, Gella, Liu, and Taslakian]{guigroundingexplicit}
Suyuchen Wang, Tianyu Zhang, Ahmed Masry, Christopher Pal, Spandana Gella, Bang Liu, and Perouz Taslakian.
\newblock Improving gui grounding with explicit position-to-coordinate mapping, 2025{\natexlab{b}}.

\bibitem[Wang et~al.(2025{\natexlab{c}})Wang, Wu, Xie, Ding, Yang, Li, Liu, Li, Dong, Chen, Wang, Zhao, Chen, Duan, Xie, Yang, Su, Yu, Huang, Liu, Zhang, Zhang, Yue, Su, Zhu, Shen, Dai, and Wang]{mmbench}
Xuehui Wang, Zhenyu Wu, JingJing Xie, Zichen Ding, Bowen Yang, Zehao Li, Zhaoyang Liu, Qingyun Li, Xuan Dong, Zhe Chen, Weiyun Wang, Xiangyu Zhao, Jixuan Chen, Haodong Duan, Tianbao Xie, Chenyu Yang, Shiqian Su, Yue Yu, Yuan Huang, Yiqian Liu, Xiao Zhang, Yanting Zhang, Xiangyu Yue, Weijie Su, Xizhou Zhu, Wei Shen, Jifeng Dai, and Wenhai Wang.
\newblock Mmbench-gui: Hierarchical multi-platform evaluation framework for gui agents, 2025{\natexlab{c}}.

\bibitem[Wu et~al.(2025)Wu, Cheng, Yang, Zhang, Yang, Jiang, Mu, Peng, Qiao, Tan, Qin, Liden, Lin, Zhang, Zhang, Zhang, Zhang, and Gao]{guiactor}
Qianhui Wu, Kanzhi Cheng, Rui Yang, Chaoyun Zhang, Jianwei Yang, Huiqiang Jiang, Jian Mu, Baolin Peng, Bo Qiao, Reuben Tan, Si Qin, Lars Liden, Qingwei Lin, Huan Zhang, Tong Zhang, Jianbing Zhang, Dongmei Zhang, and Jianfeng Gao.
\newblock Gui-actor: Coordinate-free visual grounding for gui agents, 2025.

\bibitem[Wu et~al.(2024)Wu, Wu, Xu, Wang, Sun, Jia, Cheng, Ding, Chen, Liang, and Qiao]{osatlas}
Zhiyong Wu, Zhenyu Wu, Fangzhi Xu, Yian Wang, Qiushi Sun, Chengyou Jia, Kanzhi Cheng, Zichen Ding, Liheng Chen, Paul~Pu Liang, and Yu Qiao.
\newblock Os-atlas: A foundation action model for generalist gui agents, 2024.

\bibitem[Xie et~al.(2024)Xie, Zhang, Chen, Li, Zhao, Cao, Hua, Cheng, Shin, Lei, Liu, Xu, Zhou, Savarese, Xiong, Zhong, and Yu]{osworld}
Tianbao Xie, Danyang Zhang, Jixuan Chen, Xiaochuan Li, Siheng Zhao, Ruisheng Cao, Toh~Jing Hua, Zhoujun Cheng, Dongchan Shin, Fangyu Lei, Yitao Liu, Yiheng Xu, Shuyan Zhou, Silvio Savarese, Caiming Xiong, Victor Zhong, and Tao Yu.
\newblock Osworld: Benchmarking multimodal agents for open-ended tasks in real computer environments, 2024.

\bibitem[Xie et~al.(2025)Xie, Deng, Li, Yang, Wu, Chen, Hu, Wang, Xu, Wang, Xu, Wang, Sahoo, Yu, and Xiong]{osworldg}
Tianbao Xie, Jiaqi Deng, Xiaochuan Li, Junlin Yang, Haoyuan Wu, Jixuan Chen, Wenjing Hu, Xinyuan Wang, Yuhui Xu, Zekun Wang, Yiheng Xu, Junli Wang, Doyen Sahoo, Tao Yu, and Caiming Xiong.
\newblock Scaling computer-use grounding via user interface decomposition and synthesis.
\newblock In \emph{Conference on Neural Information Processing Systems}, 2025.

\bibitem[Xing et~al.(2025)Xing, Huang, Dong, Lu, Zhang, Zang, Cao, He, Wang, Wu, and Lin]{pyramiddrop}
Long Xing, Qidong Huang, Xiaoyi Dong, Jiajie Lu, Pan Zhang, Yuhang Zang, Yuhang Cao, Conghui He, Jiaqi Wang, Feng Wu, and Dahua Lin.
\newblock Pyramiddrop: Accelerating your large vision-language models via pyramid visual redundancy reduction.
\newblock In \emph{Conference on Computer Vision and Pattern Recognition}, 2025.

\bibitem[Xu et~al.(2025{\natexlab{a}})Xu, Chen, Wang, and Liu]{attentiondriven}
Hai-Ming Xu, Qi Chen, Lei Wang, and Lingqiao Liu.
\newblock Attention-driven gui grounding: Leveraging pretrained multimodal large language models without fine-tuning.
\newblock In \emph{The 39th Annual AAAI Conference on Artificial Intelligence}, 2025{\natexlab{a}}.

\bibitem[Xu et~al.(2025{\natexlab{b}})Xu, Wang, Wang, Lu, Xie, Saha, Sahoo, Yu, and Xiong]{aguvis}
Yiheng Xu, Zekun Wang, Junli Wang, Dunjie Lu, Tianbao Xie, Amrita Saha, Doyen Sahoo, Tao Yu, and Caiming Xiong.
\newblock Aguvis: Unified pure vision agents for autonomous gui interaction.
\newblock In \emph{International Conference on Machine Learning}, 2025{\natexlab{b}}.

\bibitem[Yang et~al.(2025{\natexlab{a}})Yang, Li, Dai, Yang, Luo, Zhao, Hu, Huang, Saha, Chen, Xu, Pan, Savarese, Xiong, and Li]{gta1}
Yan Yang, Dongxu Li, Yutong Dai, Yuhao Yang, Ziyang Luo, Zirui Zhao, Zhiyuan Hu, Junzhe Huang, Amrita Saha, Zeyuan Chen, Ran Xu, Liyuan Pan, Silvio Savarese, Caiming Xiong, and Junnan Li.
\newblock Gta1: Gui test-time scaling agent, 2025{\natexlab{a}}.

\bibitem[Yang et~al.(2025{\natexlab{b}})Yang, Wang, Li, Luo, Chen, Huang, and Li]{ariaui}
Yuhao Yang, Yue Wang, Dongxu Li, Ziyang Luo, Bei Chen, Chao Huang, and Junnan Li.
\newblock Aria-ui: Visual grounding for gui instructions.
\newblock In \emph{Proceedings of Annual Meeting of the Association for Computational Linguistics}, 2025{\natexlab{b}}.

\bibitem[Ye et~al.(2025)Ye, Li, Dai, Liu, Chen, Han, Min, Ren, Zhang, Yang, and Jin]{guiarp}
Xianhang Ye, Yiqing Li, Wei Dai, Miancan Liu, Ziyuan Chen, Zhangye Han, Hongbo Min, Jinkui Ren, Xiantao Zhang, Wen Yang, and Zhi Jin.
\newblock Gui-arp: Enhancing grounding with adaptive region perception for gui agents, 2025.

\bibitem[Yuan et~al.(2025)Yuan, Zhang, Li, Cai, Yao, Chen, Wang, Hou, Chen, Jiang, and Li]{segui}
Xinbin Yuan, Jian Zhang, Kaixin Li, Zhuoxuan Cai, Lujian Yao, Jie Chen, Enguang Wang, Qibin Hou, Jinwei Chen, Peng-Tao Jiang, and Bo Li.
\newblock Enhancing visual grounding for gui agents via self-evolutionary reinforcement learning.
\newblock In \emph{Conference on Neural Information Processing Systems}, 2025.

\bibitem[Zhang et~al.(2025{\natexlab{a}})Zhang, He, Qian, Li, Li, Qin, Kang, Ma, Liu, Lin, Rajmohan, Zhang, and Zhang]{brainedgui}
Chaoyun Zhang, Shilin He, Jiaxu Qian, Bowen Li, Liqun Li, Si Qin, Yu Kang, Minghua Ma, Guyue Liu, Qingwei Lin, Saravan Rajmohan, Dongmei Zhang, and Qi Zhang.
\newblock Large language model-brained gui agents: A survey, 2025{\natexlab{a}}.

\bibitem[Zhang et~al.(2025{\natexlab{b}})Zhang, Xu, Zhu, Dai, Qiu, Yang, Luo, Chen, Wagle, Franklin, and Guo]{phiground}
Miaosen Zhang, Ziqiang Xu, Jialiang Zhu, Qi Dai, Kai Qiu, Yifan Yang, Chong Luo, Tianyi Chen, Justin Wagle, Tim Franklin, and Baining Guo.
\newblock Phi-ground tech report: Advancing perception in gui grounding, 2025{\natexlab{b}}.

\bibitem[Zhang et~al.(2025{\natexlab{c}})Zhang, Fu, Zhang, Yang, Du, Xi, Wang, Huang, Qin, Luo, and Luan]{hyperclick}
Shaojie Zhang, Pei Fu, Ruoceng Zhang, Jiahui Yang, Anan Du, Xiuwen Xi, Shaokang Wang, Ying Huang, Bin Qin, Zhenbo Luo, and Jian Luan.
\newblock Hyperclick: Advancing reliable gui grounding via uncertainty calibration, 2025{\natexlab{c}}.

\bibitem[Zhang et~al.(2025{\natexlab{d}})Zhang, Lu, Wang, Rao, Yang, and Zhu]{flexselect}
Yunzhu Zhang, Yu Lu, Tianyi Wang, Fengyun Rao, Yi Yang, and Linchao Zhu.
\newblock Flexselect: Flexible token selection for efficient long video understanding.
\newblock In \emph{Conference on Neural Information Processing Systems}, 2025{\natexlab{d}}.

\bibitem[Zhao et~al.(2025{\natexlab{a}})Zhao, Chen, Inan, Kessler, Wang, Wutschitz, Yang, Zhang, Minervini, Rajmohan, and Sim]{guicursor}
Yu Zhao, Wei-Ning Chen, Huseyin~Atahan Inan, Samuel Kessler, Lu Wang, Lukas Wutschitz, Fangkai Yang, Chaoyun Zhang, Pasquale Minervini, Saravan Rajmohan, and Robert Sim.
\newblock Learning gui grounding with spatial reasoning from visual feedback, 2025{\natexlab{a}}.

\bibitem[Zhao et~al.(2025{\natexlab{b}})Zhao, Chen, Inan, Kessler, Wang, Wutschitz, Yang, Zhang, Minervini, Rajmohan, and Sim]{spatialgui}
Yu Zhao, Wei-Ning Chen, Huseyin~Atahan Inan, Samuel Kessler, Lu Wang, Lukas Wutschitz, Fangkai Yang, Chaoyun Zhang, Pasquale Minervini, Saravan Rajmohan, and Robert Sim.
\newblock Learning gui grounding with spatial reasoning from visual feedback, 2025{\natexlab{b}}.

\bibitem[Zheng et~al.(2024)Zheng, Gou, Kil, Sun, and Su]{gpt4twebagent}
Boyuan Zheng, Boyu Gou, Jihyung Kil, Huan Sun, and Yu Su.
\newblock Gpt-4v(ision) is a generalist web agent, if grounded.
\newblock In \emph{Forty-first International Conference on Machine Learning}, 2024.

\end{thebibliography}
}

\clearpage
\setcounter{page}{1}
\maketitlesupplementary

\section{Details About Attention Heuristic Cropping}
This section details our exploration of leveraging attention scores to better locate regions containing target UI elements in screenshots. Large Vision-Language Models (LVLMs) inherently possess strong text-visual alignment capabilities. Prior work indicates that text-to-vision attention scores from specific decoder layers can effectively locate instruction-relevant visual patches~\cite{flexselect,attentiondriven}. Furthermore, models adaptively adjust the attention assigned to visual tokens during text generation~\cite{dycoke}. Denoting the visual tokens as $V \in \mathbb{R}^{L_v \times d}$, we experiment with different text tokens as queries to compute the attention scores:

\begin{itemize}
    \item Using all instruction tokens $T_{\text{instruct}}$ as queries, averaging the final scores over the text length dimension.
    \item Using the first generated token ``\textless im\_start\textgreater'' as the query.
    \item Using the comma token from the generated coordinate format ``(x, y)'' as the query, an insight inspired by GUI-Actor~\cite{guiactor}.
    \item Using the final generated token ``\textless im\_end\textgreater'' as the query.
\end{itemize}

We conduct experiments with GTA1-7B on the ScreenSpot-Pro benchmark. Following the cropping procedure described in Section 3.1, we derive attention scores from the 20th decoder layer, set $k=100$ and $m=4$, and then evaluate two metrics: the ratio of top-$m$ regions containing the target bounding box, and the final grounding accuracy after clustering.

\begin{table}[H]
\centering
\footnotesize
\begin{tabular}{lcc}
\hline
Query Tokens & Target BBox Containing Ratio & SS-Pro Avg. \\
\hline
$T_{\text{instruct}}$ & 79.5\% & 60.5\\
$T_{\text{\textless im\_start\textgreater}}$ & 73.1\% & 52.2 \\
$T_{\text{\textless im\_end\textgreater}}$ & 50.9\% & 33.3 \\
$T_{\text{comma}}$ & \textbf{83.4\%} & \textbf{61.7} \\
\hline
\end{tabular}
\caption{Comparison of cross-attention scores computed using different query tokens. The comma token yields the best performance and is therefore chosen as our default setting.}
\label{tab:performance_comparison}
\vspace{-15pt}
\end{table}

Our results(Table~\ref{tab:performance_comparison}) show that using the comma token as the query yields the best localization performance, with 83.4\% of the 4 selected views containing the target bounding box, which also translates to the highest final grounding accuracy. Consequently, we adopt this as our default configuration.

\section{Coordinate Selection via Trained Model}
In this section, we explore an alternative to clustering: training a dedicated model to select the correct coordinate from multiple candidate predictions. The motivation stems from Figure 1(b), which shows that the probability of having at least one correct prediction among the views (Pass@N) increases with the number of views. However, as shown in Table~\ref{tab:cluster_vs_passn}, while our clustering method significantly surpasses the single-view baseline, its accuracy remains lower than the Pass@N upper bound. This indicates a potential performance gap that could be bridged by a perfect selection model.

\begin{table}[H]
\centering
\footnotesize
\begin{tabular}{lcc}
\hline
View Number & Clustering Acc & Pass@N Acc \\
\hline
2 & 61.0 & 69.0 \\
4 & 61.7 & 70.2 \\
10 & 60.6 & 73.0 \\
\hline
\end{tabular}
\caption{Comparison between clustering accuracy and Pass@N accuracy. The gap indicates the potential room for improvement with an ideal selection model.}
\label{tab:cluster_vs_passn}
\vspace{-15pt}
\end{table}

\paragraph{Data Preparation}
We utilize the open-source GUI grounding dataset from GTA1~\cite{gta1}. The data is firstly filtered with the following rules: (1) image resolution larger than $2560 \times 1440$; (2) bounding box area smaller than $500$ $\text{pixels}^2$. This process yields approximately 20k samples. For each sample, we annotate 2-4 distinct red points on the image, each with a numerical label, as shown in Figure~\ref{fig:premi}. One point is placed within the target bounding box, while the others are randomly distributed outside it. The annotation metadata, including the instruction, target bounding box, point coordinates, and the image, is saved for training.

\begin{figure}[H]
  \centering
  \includegraphics[width=0.98\linewidth]{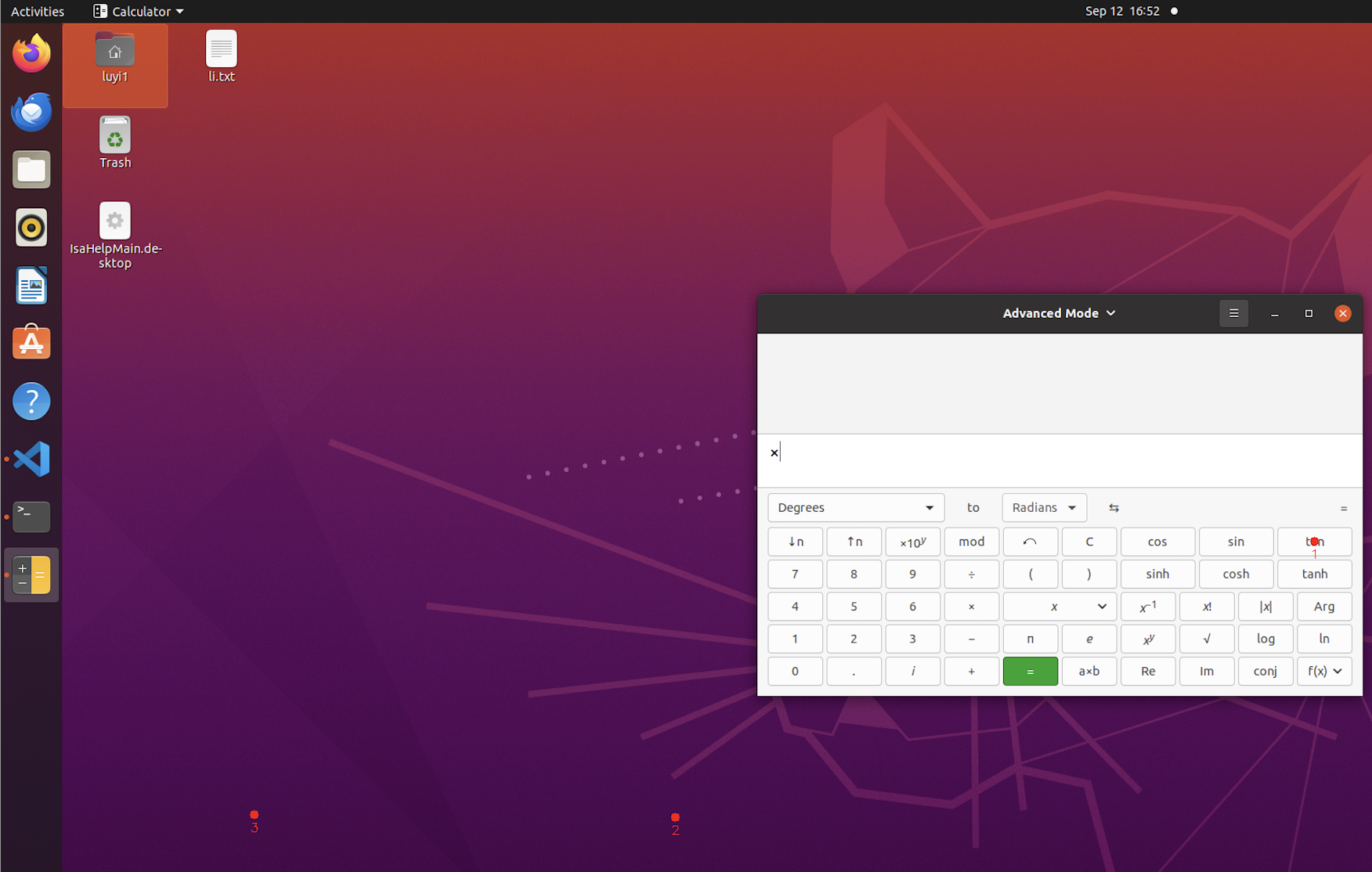}
  \caption{Example of annotated image. We annotate 2-4 visible red dots with corresponding numerical label for each sample. The model is trained to directly output the correct label.}
  \label{fig:premi}
  \vspace{-15pt}
\end{figure}
\begin{figure*}[t]
  \centering
  \begin{subfigure}{0.95\linewidth}
    \includegraphics[width=\linewidth]{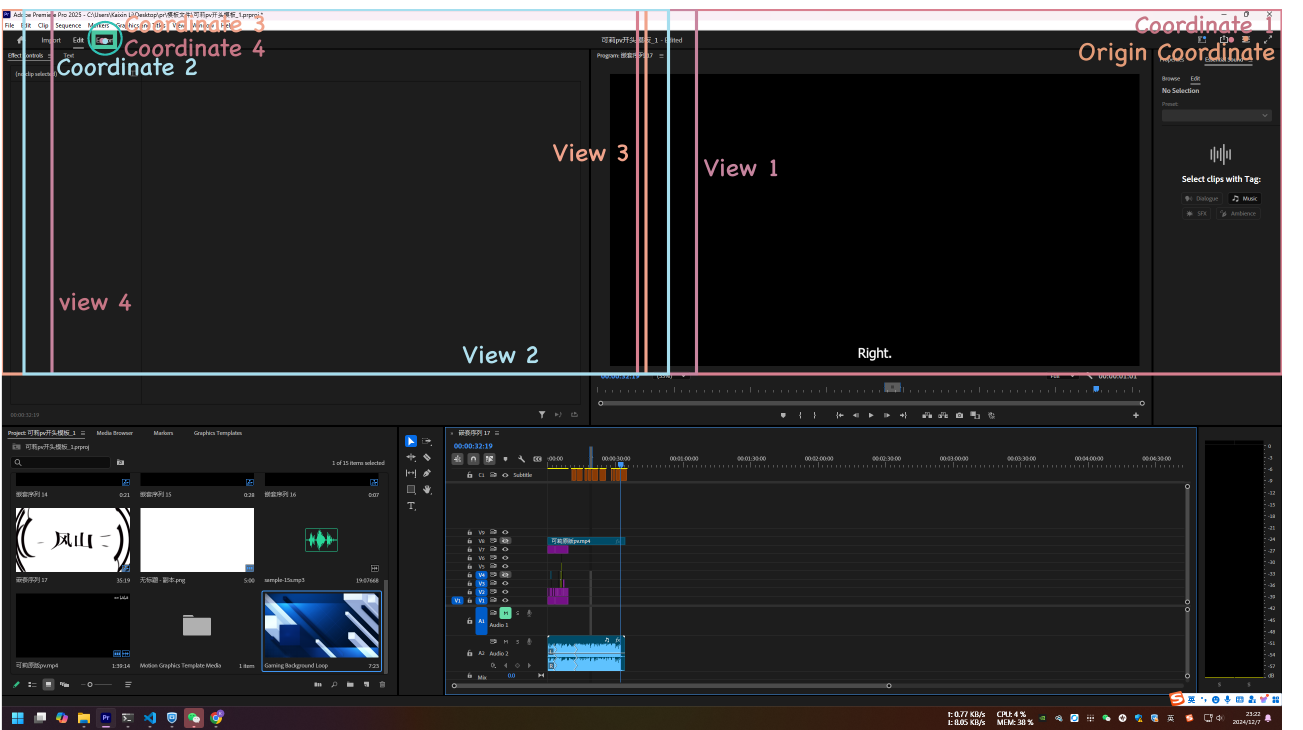}
    \label{fig:supp_example1}
  \end{subfigure}
  \vspace{-15pt}
  \caption{Multi-view example from SS-Pro evaluated by GTA1-7B. Instruction is ``change to export workspace''.}
  \vspace{-15pt}
\end{figure*}
\paragraph{Model Training}
We employ GRPO (Guided Reinforcement Policy Optimization) to train a model to directly output the numerical label of the correct point. The model takes the annotated image and user instruction as input. The rule-based reward is defined as follows: if the model outputs the correct point label, the reward is 1; otherwise, it is 0. We use Qwen3VL-4B-Instruct as the base model and train it on 8 A6000 GPUs, with 8 rollouts per group and a gradient accumulation step of 32, for a total of 170 optimization steps. The average reward converged, rising from 0.47 to 0.68.

\begin{tcblisting}{
    listing only,
    listing engine=listings,
    listing options={
        basicstyle=\footnotesize\ttfamily,
        breaklines=true,
        columns=fullflexible,
        backgroundcolor=\color{blue!5!white},
        frame=none,
        numbers=none
    },
    colback=blue!5!white,
    colframe=blue!75!black,
    title=Prompt For Coordinate Selection Model,
    fonttitle=\bfseries,
    width=\linewidth,
    listing remove caption=true
}
System Prompt:
You are an expert UI element verifier. Given a original GUI screenshot, the GUI screenshot annotated with some numbered candidate points (each marked with a red dot and a corresponding number label under the dot) and  a user's instruction, you are expected to choose single most appropriate point that user most likely to click based on the instruction step by step. Return the annotated number under the optimal point in bracket: [Number].

User Prompt:
Instruction + Annotated Image

Output Format:
[Number Label]
\end{tcblisting}
\label{fig:prompt_coordinate_selection}

\paragraph{Evaluation and Analysis}
We evaluate the trained model by having it determine the final coordinate from multiple view predictions, with the expectation that it could achieve performance close to the Pass@N upper bound. Specifically, after obtaining coordinate predictions from diverse views, we annotate them as red dots with number labels on the screenshot and prompt the trained model to generate the label of the point a user is most likely to click based on the instruction.

\begin{table}[H]
\centering
\footnotesize
\begin{tabular}{lcc}
\hline
Base Model & Aggregation Method & SS-Pro Avg. \\
\hline
GTA1-7B & Qwen3VL-4B-Instruct & 60.5 \\
GTA1-7B & Qwen3VL-4B-Instruct (Trained) & 62.8 \\
GTA1-7B & Clustering (Ours) & \textbf{61.7} \\
Qwen3VL-8B-Instruct & Qwen3VL-4B-Instruct  & 62.7 \\
Qwen3VL-8B-Instruct & Qwen3VL-4B-Instruct (Trained) & 65.3 \\
Qwen3VL-8B-Instruct & Clustering (Ours) & \textbf{65.5} \\
\hline
\end{tabular}
\caption{Performance comparison when using another LVLM versus our clustering method for coordinate aggregation. The training improves performance of selector model over it's baseline, but still fails to consistently outperform the simple clustering.}
\label{tab:selector_vs_cluster}
\vspace{-15pt}
\end{table}

As shown in Table~\ref{tab:selector_vs_cluster}, the trained selector model fails to consistently surpass our clustering method. While it shows a minor improvement, it is outperformed by clustering in the critical comparison with Qwen3VL-8B-Instruct. This result suggests that training a separate model for this selection task is not an effective strategy, as the performance gain is marginal and inconsistent, failing to justify the additional complexity and training cost. The clustering method remains a more robust and reliable aggregation strategy.

\section{Case Study}

\begin{figure*}[t]
  \centering
  \begin{subfigure}{0.95\linewidth}
    \includegraphics[width=\linewidth]{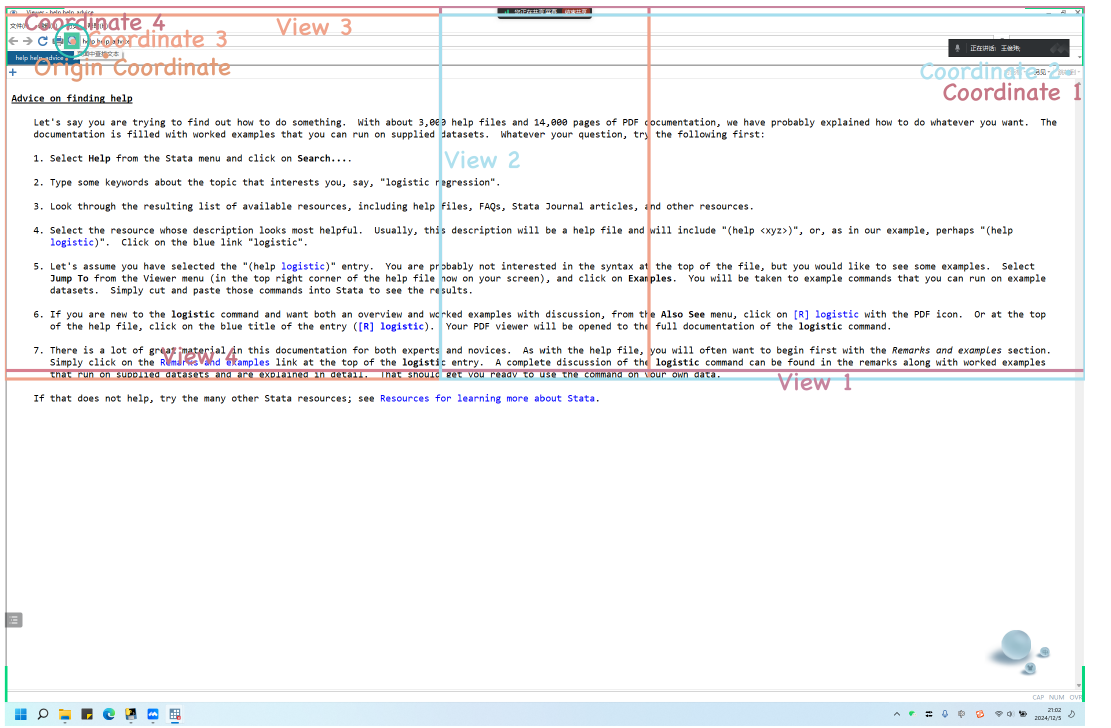}
    \label{fig:supp_example2}
  \end{subfigure}
  \vspace{-15pt}
  \caption{Multi-view example from SS-Pro evaluated by GTA1-7B. Instruction is ``find text on the page".}
\end{figure*}

\begin{figure*}[t]
  \centering
  \begin{subfigure}{0.95\linewidth}
    \includegraphics[width=\linewidth]{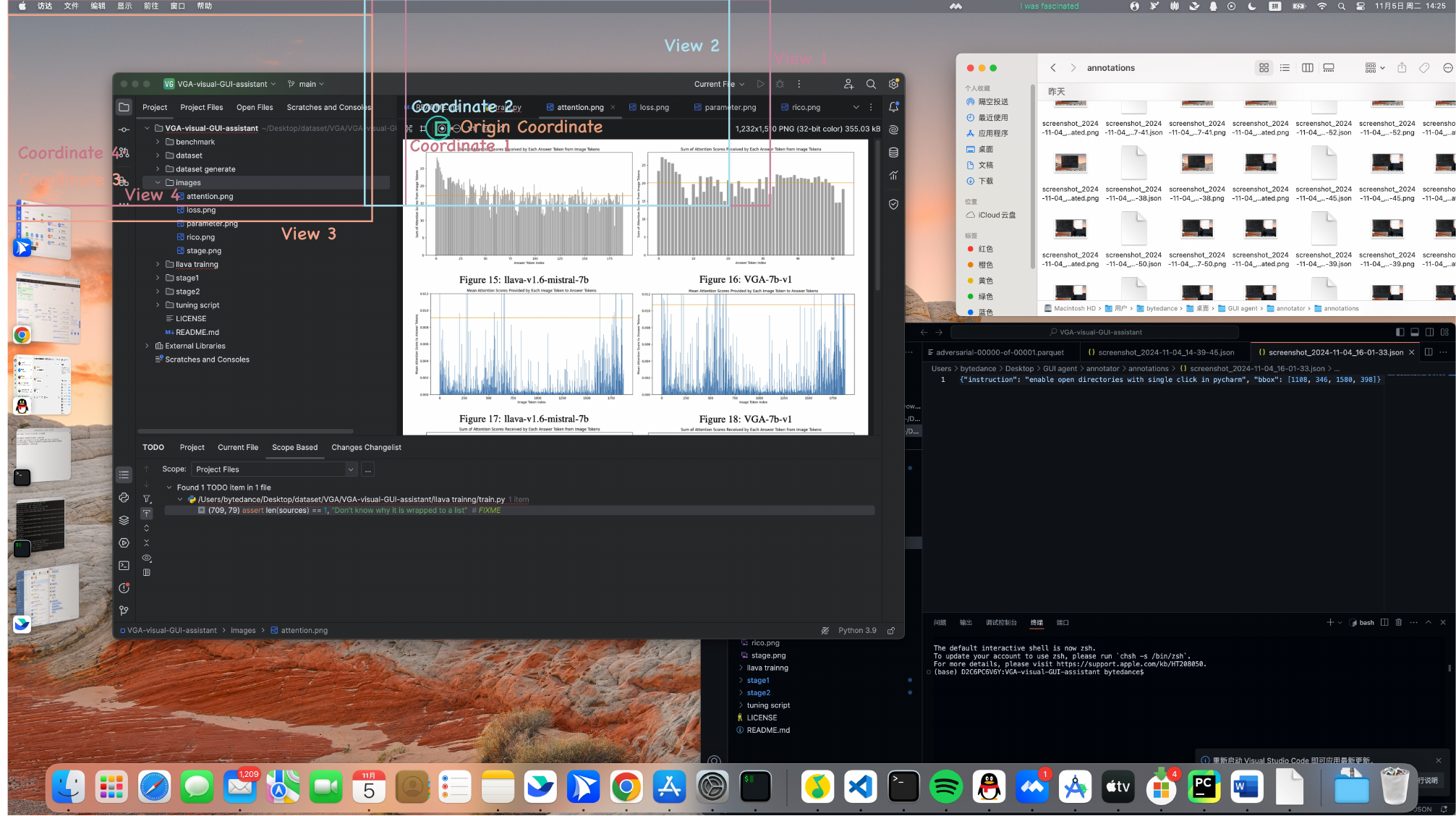}
    \label{fig:supp_example3}
  \end{subfigure}
  \vspace{-15pt}
  \caption{Multi-view example from SS-Pro  evaluated by GTA1-7B. Instruction is ``zoom in the image in pycharm".}
\end{figure*}

\end{document}